\documentclass[10pt,journal,compsoc]{IEEEtran}
\ifCLASSINFOpdf
\else
\fi
\usepackage{graphicx}
\usepackage{times}
\usepackage{epsfig}
\usepackage{amsmath}
\usepackage{bm}
\usepackage{amssymb}
\usepackage{ltablex}
\usepackage{stmaryrd}
\usepackage{mathtools}
\usepackage{multicol}
\usepackage{multirow}
\usepackage{lipsum}
\usepackage{bbm}
\usepackage{soul}
\usepackage{color}
\usepackage{adjustbox}

\usepackage{amsmath,amsfonts,bm}

\def\eqref#1{equation~\ref{#1}}

\def\1{\bm{1}}

\def\vb{{\bm{b}}}

\def\mC{{\bm{C}}}

\def\mI{{\bm{I}}}

\def\mM{{\bm{M}}}

\def\mX{{\bm{X}}}

\DeclareMathAlphabet{\mathsfit}{\encodingdefault}{\sfdefault}{m}{sl}
\SetMathAlphabet{\mathsfit}{bold}{\encodingdefault}{\sfdefault}{bx}{n}

\usepackage[backend=bibtex,bibstyle=ieee,citestyle=numeric-comp]{biblatex}
\bibliography{egbib}

\usepackage[linesnumbered,ruled,vlined]{algorithm2e}

\SetCommentSty{mycommfont}
\SetKwInput{KwData}{Input}
\SetKwInput{KwResult}{Output}

\newcommand{\RNum}[1]{\uppercase\expandafter{\romannumeral #1\relax}}
\newcommand{\revision}[1]{{{#1}}}

\DeclarePairedDelimiter{\norm}{\lVert}{\rVert} 
\usepackage{hyperref}

\usepackage[capitalize]{cleveref}
\crefname{section}{Sec.}{Secs.}
\Crefname{section}{Section}{Sections}
\Crefname{table}{Table}{Tables}
\crefname{table}{Tab.}{Tabs.}

\usepackage{siunitx}
\usepackage{booktabs}
\usepackage{etoolbox}
\renewrobustcmd{\bfseries}{\fontseries{b}\selectfont}
\renewrobustcmd{\boldmath}{}
\newrobustcmd{\B}{\bfseries}

\newcommand{\etal}{\textit{et al.}}

\begin{document}
%
\title{Structure-Guided Image Completion with Image-level and Object-level Semantic Discriminators}


\author{{Haitian Zheng$^{1,2}$} \  Zhe~Lin$^{2}$ \  Jingwan~Lu$^{2}$ \  Scott~Cohen$^{2}$ \ Eli~Shechtman$^{2}$ \ Connelly~Barnes $^{2}$ \  Jianming Zhang$^{2}$ \  Qing~Liu$^{2}$ \  Sohrab~Amirghodsi$^{2}$ \ Yuqian~Zhou$^{2}$ \  Jiebo~Luo$^{1}$\\
{$^{1}$}University of Rochester \quad {$^{2}$}Adobe Research\\
$^{1}${\tt\small \{hzheng15,jluo\}@cs.rochester.edu} \quad $^{2}${\tt\small \{hazheng,zlin,jlu,scohen,jianmzha,elishe,cbarnes,qliu,tamirgho,yzhou\}@adobe.com}
}

\author{
        Haitian~Zheng,~\IEEEmembership{Student Member,~IEEE,}
        Zhe~Lin,~\IEEEmembership{Member,~IEEE,}
        Jingwan~Lu,~\IEEEmembership{Member,~IEEE,}
        Scott~Cohen,~\IEEEmembership{Member,~IEEE,}
        Eli~Shechtman,~\IEEEmembership{Member,~IEEE,}
        Connelly~Barnes,~\IEEEmembership{Member,~IEEE,}
        Jianming~Zhang,~\IEEEmembership{Member,~IEEE,}
        Qing~Liu,~\IEEEmembership{Member,~IEEE,}
        Sohrab~Amirghodsi,~\IEEEmembership{Member,~IEEE,}
        Yuqian~Zhou, \IEEEmembership{Member,~IEEE,}
        Jiebo~Luo,~\IEEEmembership{Fellow,~IEEE,}
\IEEEcompsocitemizethanks{\IEEEcompsocthanksitem Haitian~Zheng and Jiebo~Luo are with the Department
of Computer Science, University of Rochester.\protect\\
E-mail: \{hzheng15,jluo\}@cs.rochester.edu
\IEEEcompsocthanksitem Haitian~Zheng, Zhe~Lin, Jingwan~Lu, Scott~Cohen, Eli~Shechtman, Connelly~Barnes, Jianming~Zhang, Qing~Liu, Sohrab~Amirghodsiand and Yuqian Zhou are with Adobe Research.\protect\\
E-mail: \{hazheng,zlin,jlu,scohen,jianmzha,nxu\}@adobe.com}
}

\markboth{}%
{Shell \MakeLowercase{\textit{et al.}}: Bare Demo of IEEEtran.cls for Computer Society Journals}
%



\IEEEtitleabstractindextext{%
\begin{abstract}
Structure-guided image completion aims to inpaint a local region of an image according to an input guidance map from users. While such a task enables many practical applications for interactive editing, existing methods often struggle to hallucinate realistic object instances in complex natural scenes. Such a limitation is partially due to the lack of semantic-level constraints inside the hole region as well as the lack of a mechanism to enforce realistic object generation. In this work, we propose a learning paradigm that consists of semantic discriminators and object-level discriminators for improving the generation of complex semantics and objects. Specifically, the semantic discriminators leverage pretrained visual features to improve the realism of the generated visual concepts. Moreover, the object-level discriminators take aligned instances as inputs to enforce the realism of individual objects. Our proposed scheme significantly improves the generation quality and achieves state-of-the-art results on various tasks, including segmentation-guided completion, edge-guided manipulation and panoptically-guided manipulation on Places2 datasets. Furthermore, our trained model is flexible and can support multiple editing use cases, such as object insertion, replacement, removal and standard inpainting. In particular, our trained model combined with a novel automatic image completion pipeline achieves state-of-the-art results on the standard inpainting task.
\end{abstract}

}

\maketitle

\IEEEdisplaynontitleabstractindextext

\IEEEpeerreviewmaketitle

\begin{figure*}[t]
	\centering
    \includegraphics[width=1.\linewidth]{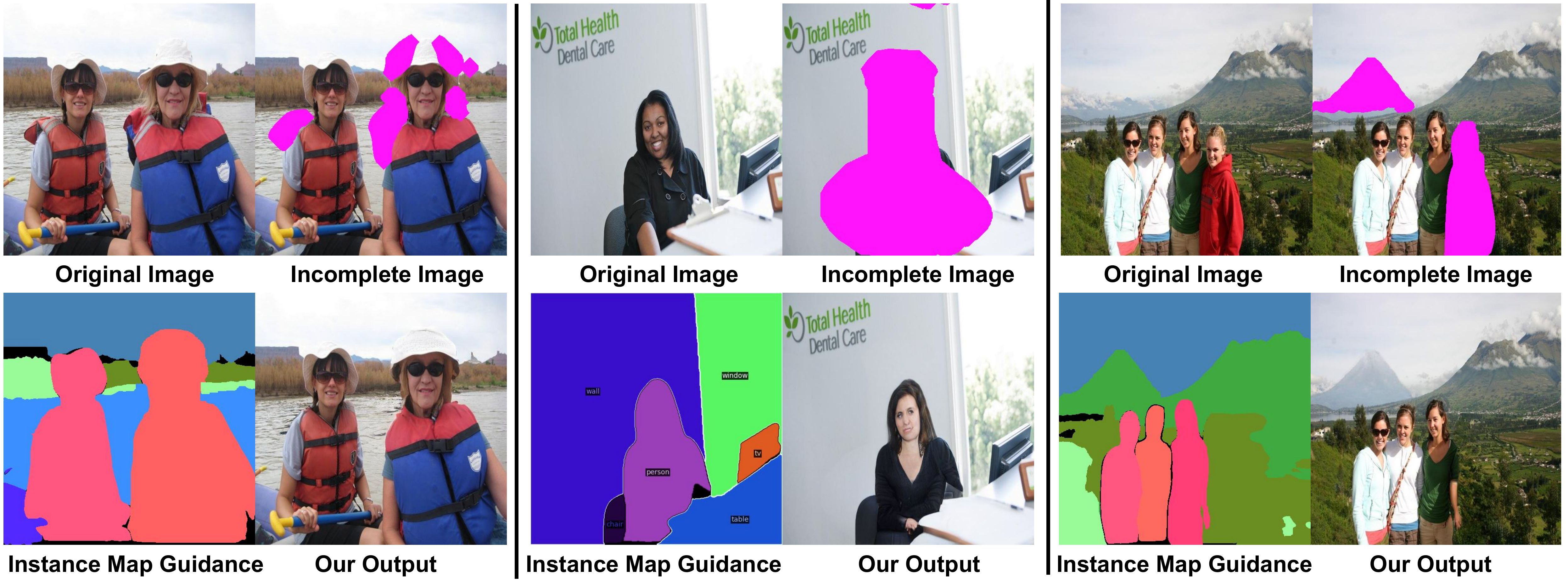}
	\caption{
	We propose a new structure-guided image completion model that leverages our proposed semantic discriminators and object-level discriminators for photo-realistic image completion.
	Our trained guided completion model enables multiple image editing applications, including local image manipulation (left), person anonymization (middle), layout manipulation (right) and instances removal (right).
	}
	\label{fig:teaser}
\end{figure*}

\section{Introduction}
Recently, there have been increasing attention to and demand on guided image completion~\cite{edgeconnect,deepfill,foreground_aware,zheng2022semantic,hong2018learning,sesame,sketchedit,faceshop,face_editing2019}
for photo editing and creative expression.
The aim of the guided image completion is to complete the missing region of an image according to an optional guidance map such as semantic label map~\cite{hong2018learning,sesame}, edge map~\cite{deepfill,deepfillv2,edgeconnect,sketchedit},
or colored pixels~\cite{faceshop,face_editing2019}.
Such tasks are shown to enable versatile image editing operations such as
completion of large missing regions~\cite{deepfill,deepfillv2,edgeconnect,lama},  object removal~\cite{profill,cmgan,sesame}, insertion~\cite{hong2018learning,qiu2020hallucinating,sesame}, replacement~\cite{zhu2017unpaired,hong2018learning,sesame} and manipulating the layout of an image~\cite{sketchedit,zheng2022semantic,sesame,edgeconnect}.


By posing guided image completion as a conditional inpainting problem~\cite{song2018spg,deepfillv2,edgeconnect,sesame,sketchedit},
image manipulation have been made significant progress in the past. However, due to the lack of a mechanism to ensure the quality of the generated objects and semantic structures, the current guided completion models are often limited in synthesizing a large missing region of complex natural scenes and randomly placed objects, leading to obvious structural artifacts such as distorted objects and degenerated semantic layout, as depicted in \cref{fig:teaser}.
As such, how to hallucinate large missing regions while maintaining a reasonable semantic layout and realistic object instances remains an open and essential problem for guided image completion.


We tackle the challenging large-hole guided completion task where the goal is to complete whole objects or large parts of objects that could be arbitrarily located in a natural scene.
Different from methods~\cite{sesame,song2018spg,edgeconnect,sketchedit,deepfillv2} that treat guided completion as a straightforward extension of inpainting~\cite{deepfill,deepfillv2,pathak2016context,partial_convolutions,comodgan,lama,cmgan} with additional conditions,
we argue that properly imposed semantic and object-level constraints on the generated image are crucial for improving the realism of complex semantic layouts and objects details.

Along this line, we propose a new learning paradigm under the framework of GANs~\cite{gan} as GANs offer flexibility on adding explicit constraints and inference efficiency compared to approaches including the recent diffusion-based methods~\cite{ddpm,iddpm,ldm}. More specifically, we proposed \emph{semantic discriminators} and \emph{object-level discriminators} to enforce the realism of generated semantic layout and objects details: our \emph{semantic discriminators} leverage the semantic understanding capacity of pretrained visual models~\cite{clip} to enhance the model capacity on semantic-level discrimination, facilitating more semantically plausible generation outcome; the \emph{object-level discriminators} take the aligned and cropped object as input to better determine the quality of fine-grained objects instances at a local scale, imposing stronger constraints to the appearance of local objects.

To facilitate editing of instances and standard inpainting where guidance map is in absence, we further propose a novel \emph{instance-guided completion task}. Different from edge~\cite{edgeconnect,deepfillv2,sketchedit} or segmentation~\cite{sesame} guided completion, panoptic labeling~\cite{panoptic} provides fine-grained and instance-level semantics information and enables instance-level image editing.
Furthermore, we proposed a \emph{fully automatic image completion pipeline} to enable standard image inpainting. Specifically, our pipeline is based on predicting the panoptic segmentation inside the missing region following an instance-guided inpainting stage to address the absence guidance information inside the hole for inpainting.

With the newly introduced semantic and object-level discriminators and the new standard inpainting pipeline, our method significantly boosts the realism of the completed objects and leads to significant gain on various inpainting tasks including segmentation-guided image inpainting, edge-guided image inpainting, instance-guided image inpainting and standard image inpainting without guidance, producing very promising results for large-hole image completion and object completion.

In summary, our contributions are three-fold:
\begin{itemize}
    \item A new semantic discriminator design that leverages the pretrained visual features to encourage the semantic consistency of the generated contents and a novel object-level discriminator framework that enforces the realism of the generated local objects for guided image completion.
    \item State-of-the-art results on the Places2 dataset for various tasks including segmentation-guided inpainting, edge-guided inpainting, and instance-guided inpainting compared to other methods.
    \item Flexibility of our trained model on multiple image editing use cases, such as object manipulation, replacement, removal and standard inpainting. Combined with an automatic image completion pipeline, our trained model achieves the state-of-the-art results on the standard inpainting task.
    
\end{itemize}

\begin{figure*}[h!]
	\centering
	\includegraphics[width=0.95\linewidth]{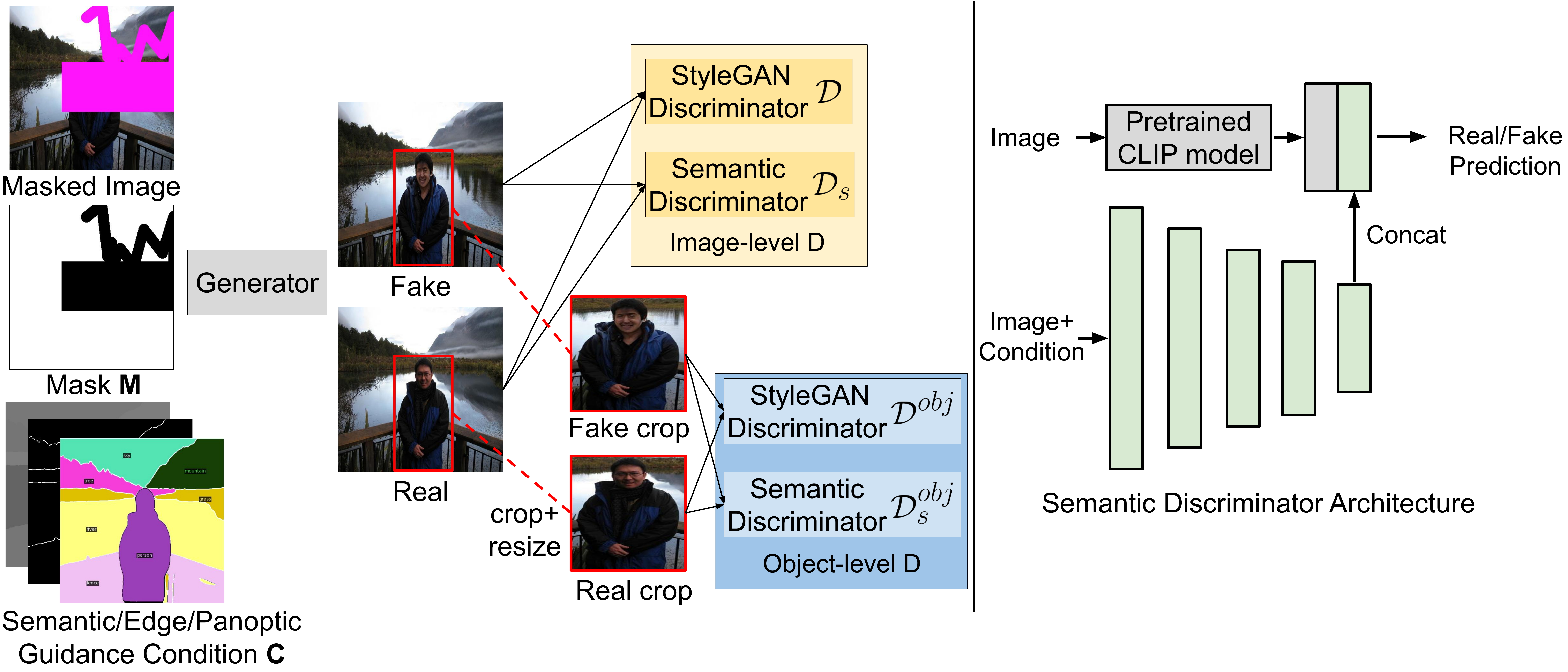}
	\caption{
		\small{
			Left: Our model can take an edge/segmentation/panoptic map as condition for guided image completion and leverages a combination of vanilla StyleGAN discriminator~\cite{stylegan2} and the proposed semantic discriminators at both image level and object level to enforce semantic and object coherency.
			The object-level discriminators take the resized object crop as inputs to enforce realism of object instances. Right: the semantic discriminators leverage the semantic knowledge of the pretrained CLIP~\cite{clip} model to enforce the realism of generated semantic.
		}
	}
	\label{fig:pipeline}
\end{figure*}

\section{Related work}
\subsection{Image Inpainting}
Image inpainting has a long-standing history.
Early approaches leverage patch-based copy-pasting~\cite{Image_quilting,kwatra2005texture,patchmatch,cho2008patch,image_melding} or color propagation~\cite{ballester2001filling,chan2001nontexture,shen2002mathematical,criminisi2004region} to fill in the target hole. Those methods can produce high-quality textures while completing simple shapes cannot complete new semantic structures due to the lack of mechanism for modeling the high-level semantic information.
With the development of deep generative models, image inpainting has gradually shifted to data-driven paradigm, where generative models are trained to predict the contents inside the missing region.
By training deep generative models~\cite{gan,esser2021taming} on large scale datasets, the learned models can capture higher-level representation of images and hence can generate more visually plausible results.
Specifically, \cite{pathak2016context} train an encoder-decoder network that completes the missing region of an image for hole filling.
Later, numerous approaches have been proposed to improve the learning-based holl filling. Motivate to decompose structure and texture~\cite{aujol2006structure,bertalmio2003simultaneous,liu2022delving}, two-stage networks are proposed to predict smoothed image~\cite{deepfill,MEDFE,hifill,ict,profill}, edge~\cite{edgeconnect,foreground_aware,gao2023macroscopic,qian2023adaptive}, gradient~\cite{yang2020learning} or segmentation map~\cite{song2018spg} for enhancing the final output.

Attention mechanism~\cite{deepfill}, dilated convolution~\cite{iizuka2017globally,deepfill} and Fourier convolution~\cite{chi2020fast} are are incorporated to expand the receptive field of the generative models, allowing the model to better propagate contextual features to the missing region.
Moreover, feature gating mechanism such as partial convolution~\cite{partial_conv} and gated convolution~\cite{deepfillv2} is proposed to filter invalid features inside the hole.
Furthermore, co-modulation~\cite{comodgan} and cascaded-modulation~\cite{cmgan} mechanisms are proposed to enhance the global prediction capacity.
Approaches based on probabilistic diffusion models~\cite{ho2020denoising,saharia2021palette,lugmayr2022repaint} and vision transformers~\cite{ict,zheng2021pluralistic,zheng2021tfill} also have shown promising results on image inpainting.

\subsection{Guided Image Inpainting}
Despite showing impressive results on completing background or partial objects, image inpainting methods would often lead to obvious structural artifacts for large holes, as learning the complex semantic structure of natural images is inherently challenging.

To facilitate more structurally realistic prediction while providing tools for users to manipulate the inpainting outcome, guided inpainting leverage additional structural guidance to improve image completion.
Specifically, edge~\cite{deepfillv2,edgeconnect,foreground_aware}, color~\cite{faceshop,face_editing2019} or gradient maps~\cite{yang2020learning} are used to guide the image completion process.
To to control the semantic layout of the inpainting results, 
semantic segmentation is also used to guide the inpainting process. Specifically, Song \etal~\cite{song2018spg} predict semantic layout to improve the realism of the generated layout. Hong \etal~\cite{hong2018learning} leverage semantic segmentation to manipulate the generated contents, Ntavelis \etal~\cite{sesame} leverage a local semantic layout to control the inpainting outcome for scene manipulation. Recently, Zhang et al.~\cite{controlnet} introduced ControNet, a method that utilizes specified input modalities to manipulate the generation results of a pre-trained text-to-image diffusion model. This method can also be applied to a pre-trained inpainting diffusion model for guided image inpainting. However, the original paper did not comprehensively evaluate guided inpainting using ControNet. Furthermore, diffusion models usually require iterative sampling and are computationally expensive.



\subsection{Discriminators for GANs}
The initial work of Generative Adversarial Networks (GANs)~\cite{gan} leverages a multilayer perceptron (MLP) to predict the realism of the generated images. 
Since then, there have been rapid progress to achieve photo-realistic image synthesis.
Specifically, patch discriminator~\cite{isola2017image,li2016precomputed,deepfillv2} is proposed to predict the realism of the generated patches in local regions.
Later, the StyleGAN-based discriminators~\cite{stylegan,stylegan2} leverages strided convolution combined with a fully connected layer to generate realistic images. 
Motivated by contrasitive learning~\cite{chen2020simple}, several works~\cite{kang2020contragan,zhang2021cross,jeong2021training,yu2021dual} design discriminators to predict the pairwise relations between different modalities or the real and fake samples. 
Likewise, the patch co-occurrence discriminator~\cite{park2020swapping} predicts the similarity between the output patches and the reference style image.
To enhance the recognition capacity of a discriminator, recent works~\cite{sauer2021projected,ensembling-gan} leverages pretrained visual features to enhance the semantic understanding of a discriminator for unconditional generation.

\section{Method}\label{sec:methodology}

\subsection{Network Architecture}
As depicted in \cref{fig:pipeline}, our guided completion model is based on conditional Generative Adversarial Networks~\cite{cgan} to complete the missing region of an image $\mX$ annotated by a binary mask $\mM$ according to a guidance map condition $\mC$ where the guidance input map can be an edge map~\cite{deepfillv2,edgeconnect,foreground_aware} or segmentation map~\cite{song2018spg,sesame} or panoptic instance labeling map~\cite{panoptic}. More details on the guidance map format can be found in the supplementary material.


\subsubsection{The Generator}\label{sec:G}
Recently, Cascaded-Modulation GAN (CM-GAN)~\cite{cmgan} has shown significant improvement in  standard image inpainting tasks thanks to the architecture design that cascades modulation blocks for better global context modeling.
Therefore, we adopt the CM-GAN generator to our guided completion task to leverage the strong inpainting capacity of the CM-GAN generator.
However, we concatenate the guidance input $\mC$ to our generator $\mathcal{G}$ so that the generator can benefit from the additional guidance. 

\begin{figure}[t]
	\centering
        \includegraphics[width=1\linewidth]{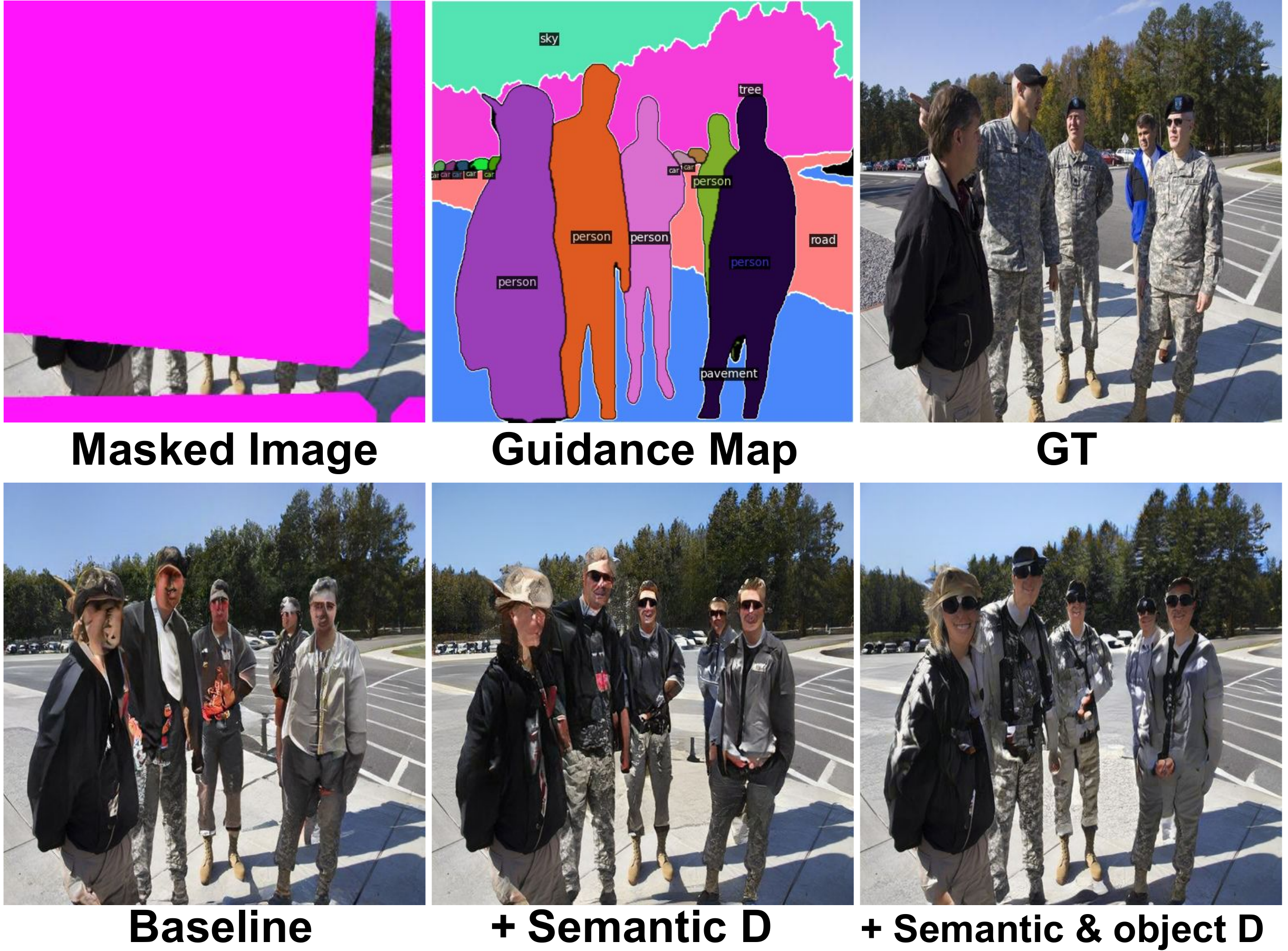}
	\caption{
		\small{
			The image-level semantic discriminator and the object-level discriminators progressively improve the photo realism of the generated image (e.g., face and body) on a guided inpainting task in comparison to the baseline trained with only the StyleGAN discriminator~\cite{stylegan2}.
		}
	}
	\label{fig:simple_ablation}
\end{figure}

\subsubsection{The Semantic and Object Discriminators}\label{sec:D_semantic}
Following recent inpainting works \cite{cmgan,comodgan} that leverage StyleGAN discriminator~\cite{stylegan2} for adversarial learning, we adopt a conditional discriminator $\mathcal{D}$ that takes the concatenation of generated image $\hat{\mX}$ and the condition $\mM,\,\mC$ as inputs to output the discriminator score $\hat{y}$:
\begin{align}
\label{eq:D}
\begin{aligned}
\hat{y} = \mathcal{D}(\hat{\mX},\,\mM,\,\mC).
\end{aligned}
\end{align}
We found that such an adversarial learning scheme indeed achieves leading results comparing to other baseline models.
However, due to the lack of further instance-level supervision and constraints on objects, the generator trained with the conditioned StyleGAN discriminator tends to hallucinate distorted objects or degenerated semantic layout as depicted in \cref{fig:simple_ablation}, which seriously impact the inpainting quality. Hence, we propose a novel \emph{semantic discriminator} for improving semantic coherency of completion and  \emph{object-level discriminators} for enhancing the photo realism of the individually generated objects.

\noindent \textbf{Semantic Discriminator.} \quad 
To generate realistic object instances and a complex semantic layout, a discriminator should distinguish whether the generated contents $\hat{\mX}$ is realistic and conformed to the given semantic layout. However, it has been shown  that~\cite{ensembling-gan,wang2020cnn} discriminators may potentially focus on artifacts that are imperceptible to humans but obvious to a classifier and that the learned visual feature may cover only parts of the visual concept~\cite{projected-gan} while ignoring other parts.
Therefore, with the regular adversarial learning between $\mathcal{G}$ and $\mathcal{D}$, it is challenging for the generator to discover complex semantic concepts or hallucinate realistic objects.

To tackle this issue, we propose a semantic discriminator $\mathcal{D}_{s}$ that leverages the visual representation extracted by a pre-trained vision model~\cite{clip} to discriminate the instance-level realism. Benefiting from the comprehensive semantic concepts captured by pre-trained vision models~\cite{bau2020understanding}, our semantic discriminator can better capture high-level visual concepts and improve the realism of the generated semantic layout, c.f. \cref{fig:simple_ablation}. Specifically, our semantic discriminator $\mathcal{D}_{s}$ takes the generated image and the panoptic condition as inputs:
\begin{align}
\label{eq:D_semantic}
\begin{aligned}
    \hat{y}_s = \mathcal{D}_{s}(\hat{\mX},\,\mM,\,\mC),
\end{aligned}
\end{align}
and output the instance-level realism prediction.
As shown in \cref{fig:pipeline} (right), the semantic discriminator is based on the two branches of the encoder to extract complementary features: a pre-trained ViT model branch~\cite{clip} (at the top) produces pre-trained semantic feature of the completed image, and a trainable encoder branch based on strided convolution (at the bottom) extracts discriminative features from the concatenation of image and condition $\hat{\mX},\mM,\mC$.
Finally, the pre-trained feature and the encoder feature at the final scale are concatenated to produce the final discriminator prediction.
As the semantic discriminator are designed to classify the high-level structure at the instance level, we found that combining the semantic discriminator $\mathcal{D}_{s}$ with the StyleGAN discriminator~\cite{ensembling-gan} $\mathcal{D}$ improves the generated local textures.

\noindent \textbf{Object-level Discriminators.} \quad 
Recent progress on image generation~\cite{stylegan,stylegan2} demonstrates impressive results on generating objects such as face, car, animal~\cite{karras2017progressive} or body~\cite{ma2017pose} in an aligned setting where objects are carefully placed or registered in the center of the image. However, generating unaligned objects in a complex natural scene is known to be challenging~\cite{styleganxl} for various tasks including inpainting~\cite{spade,comodgan} and semantic image generation~\cite{spade}. Although the semantic discriminator can improve the quality of the generated objects, generating photo-realistic instances is still challenging.
To improve the realism of completed objects, we found that the object-level alignment mechanism for the discriminator has a profound impact on improving inpainting quality. Consequently, we propose novel object-level discriminators that are dedicated to modeling the hierarchical composition of aligned objects for predicting the object-level realism. 
In particular, as shown in \cref{fig:pipeline}, given an object instance and its bounding box $\vb=(x_0,y_0,x_1,y_1)$, an object-level StyleGAN discriminator $\mathcal{D}^{obj}$ takes the crop-and-resized image $\hat{\mX}_c$ and the corresponding crop-and-resized condition maps $\mM_{c}, \mC_{c}$ as inputs to predict the realism of the object:
\begin{align}
\label{eq:D_obj}
\begin{aligned}
    \hat{y}^{obj} = \mathcal{D}^{obj}(\hat{\mX}_{c},\,\mM_{c},\,\mC_{c},\,\mI_{c}),
\end{aligned}
\end{align}
where $\mI_{c}$ is an additional binary map that indicates the shape of cropped the instance and $\hat{y}^{obj}$ represents how likely the object instance is the ground-truth object patch.

To further enhance the capacity of discriminator for object modeling, 
we also propose semantic discriminators at object-level $\mathcal{D}_{s}^{obj}$. Specifically, the discriminator $\mathcal{D}_{s}^{obj}$ takes $\hat{\mX}_{c},\,\mM_{c},\,\mC_{c},\,\mI_{c}$ as input and outputs $\hat{y}_s^{obj}$, i.e. how likely the object instance is the ground-truth object patch:
\begin{align}
\label{eq:D_obj_semantic}
\begin{aligned}
    \hat{y}_s^{obj} = \mathcal{D}_{s}^{obj}(\hat{\mX}_{c},\,\mM_{c},\,\mC_{c},\,\mI_{c}).
\end{aligned}
\end{align}
Following the architecture design of global discriminators, the object-level discriminator $\mathcal{D}^{obj}$ follows the StyleGAN discriminator~\cite{stylegan2} while the object-level semantic discriminator $\mathcal{D}_s^{obj}$ follows the implementation of image-level semantic discriminator $\mathcal{D}_s$.

\begin{figure*}[t]
	\centering
	\includegraphics[width=1\linewidth]{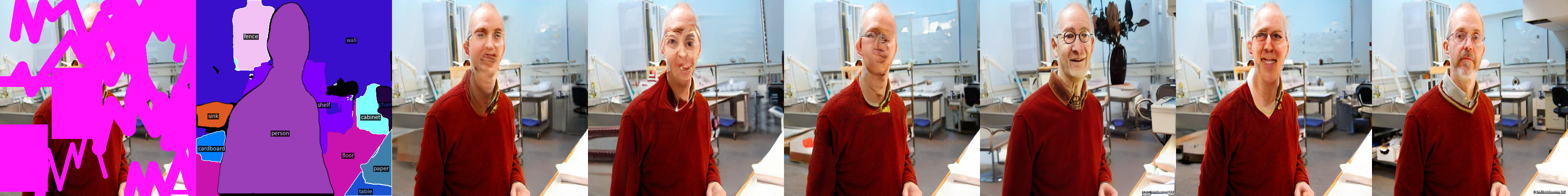}\\
	\includegraphics[width=1\linewidth]{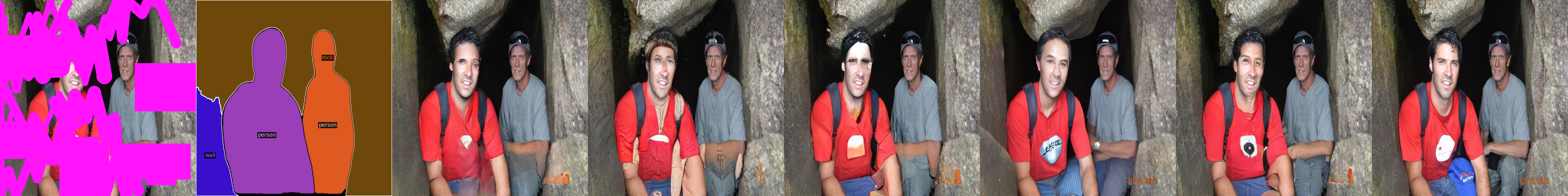}\\
	\includegraphics[width=1\linewidth]{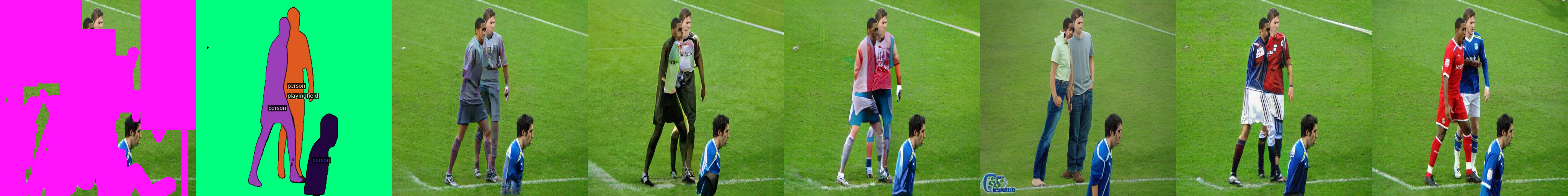}\\
	\includegraphics[width=1\linewidth]{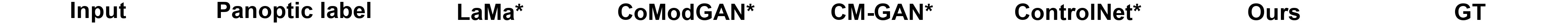}
	\caption{
		Qualitative comparisons on the guided inpainting task on Places2-person. We compare our model against LaMa*~\cite{lama}, CoModGAN*~\cite{comodgan}, CM-GAN*~\cite{cmgan} and ControlNet*~\cite{controlnet} whereas  $*$  denotes models re-trained with the additional panoptic instance segmentation condition. Best viewed by zoom-in on screen.
	}
	\label{fig:main_panoptic}
\end{figure*}

\begin{figure*}[t]
	\centering
	\includegraphics[width=1\linewidth]{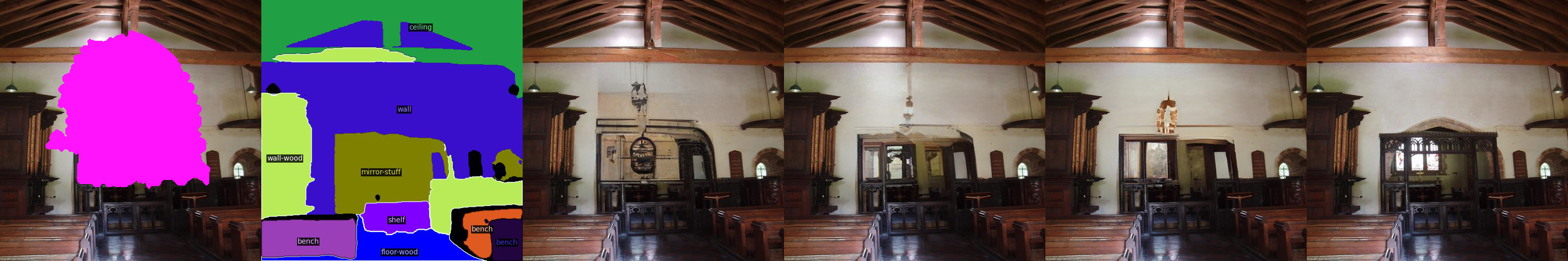}\\
	\includegraphics[width=1\linewidth]{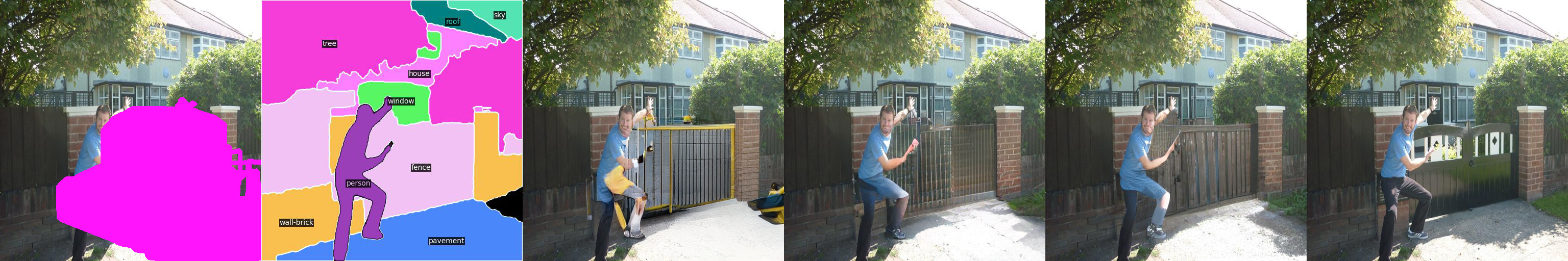}\\
	\includegraphics[width=1\linewidth]{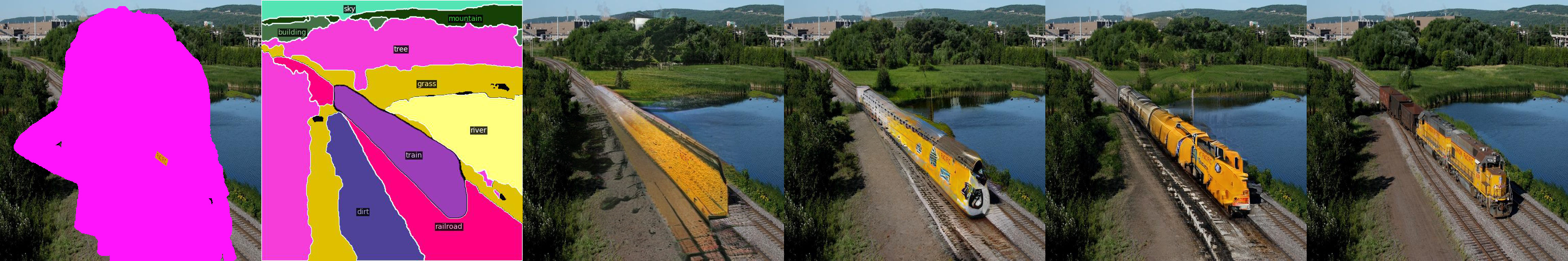}\\
	\includegraphics[width=1\linewidth]{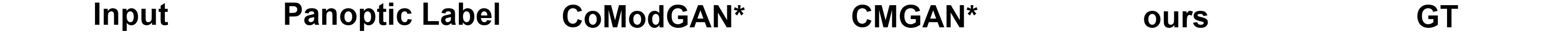}
	\caption{
		Qualitative comparisons on the  guided inpainting task on Places2-object. We compare our model against retrained CoModGAN$*$~\cite{comodgan}, CM-GAN$*$~\cite{cmgan} whereas the $*$ symbol denotes models re-trained with the additional panoptic instance segmentation condition for  guided inpainting. Best viewed by zoom-in on screen.
	}
	\label{fig:main_panoptic_object}
\end{figure*}

\begin{figure*}[t]
	\centering
	\includegraphics[width=1\linewidth]{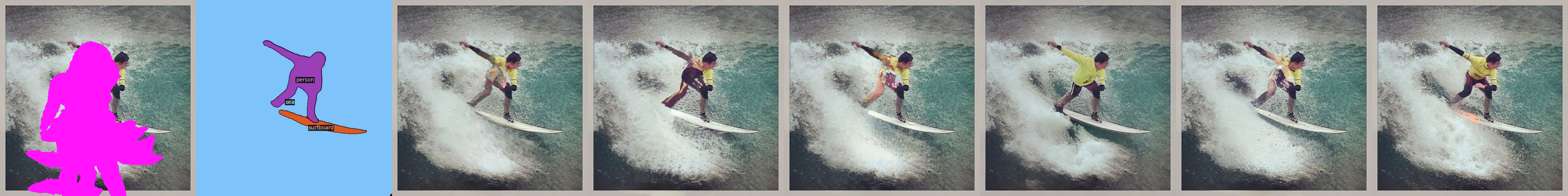}\\
	\includegraphics[width=1\linewidth]{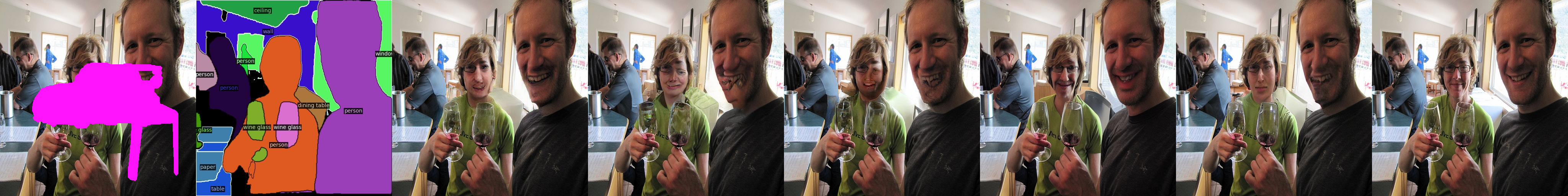}\\
	\includegraphics[width=1\linewidth]{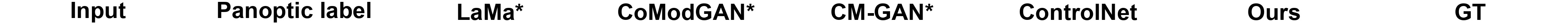}
	\includegraphics[width=1\linewidth]{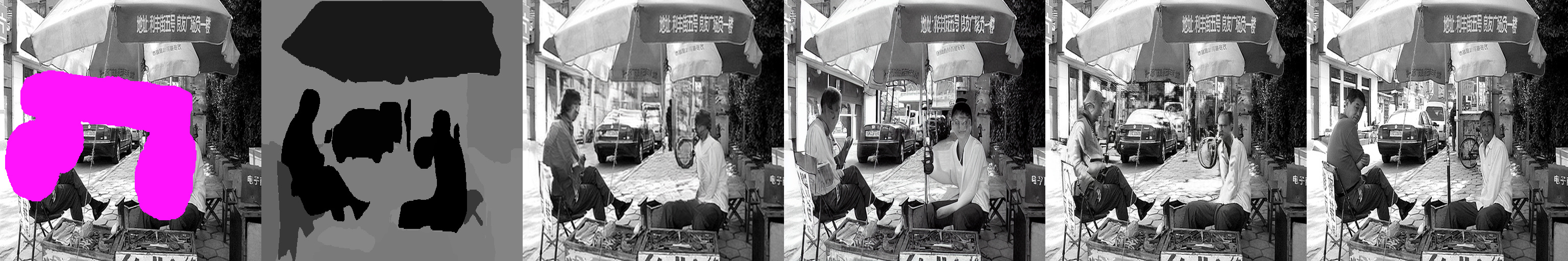}\\
	\includegraphics[width=1\linewidth]{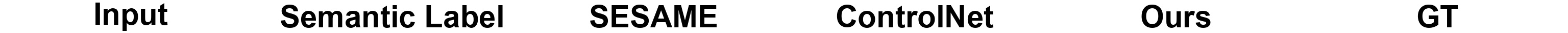}
	\includegraphics[width=1\linewidth]{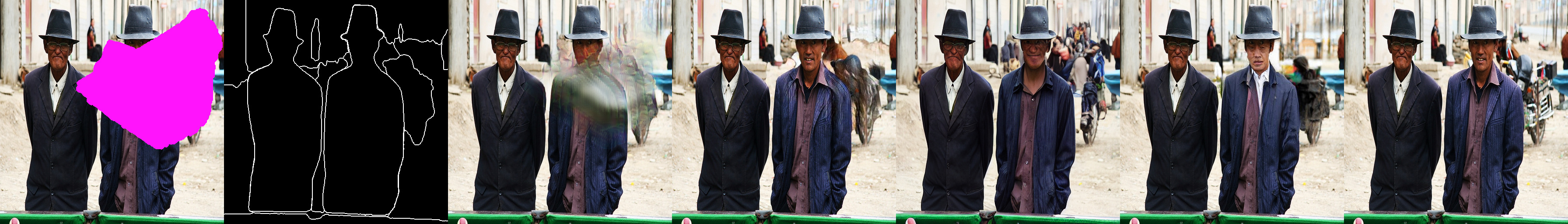}\\
	\includegraphics[width=1\linewidth]{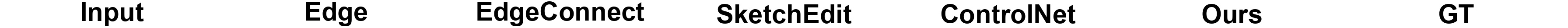}
	\includegraphics[width=1\linewidth]{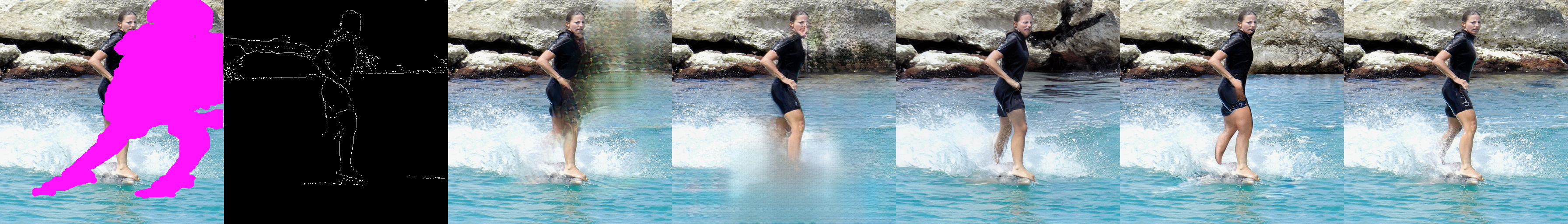}\\
	\includegraphics[width=1\linewidth]{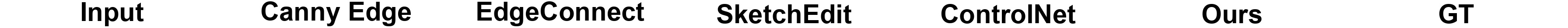}
	\caption{
		Qualitative comparisons on the guided inpainting tasks on the COCO-Stuff dataset. From top to bottom are instance-segmentation guided inpainting, semantic segmentation-guided inpainitng, instance edge guided inpaiting and Canny edge-guided inpainting. Best viewed by zoom-in on screen.
	}
	\label{fig:main_coco}
\end{figure*}

\subsubsection{Training Objective}\label{sec:objective}
Our training objective is a summation of non-saturating adversarial loss~\cite{gan} for a set of StyleGAN and semantic discriminators at both image level and object level $\mathbf{D} = \{\mathcal{D}, \mathcal{D}_s, \mathcal{D}^{obj}, \mathcal{D}_s^{obj}\}$ and perceptual loss~\cite{johnson2016perceptual} for the generator:
\begin{align}
\label{eq:adv_loss}
\begin{aligned}
    \mathcal{L} =& \sum_{\mathcal{D} \in \mathbf{D}} \log \mathcal{D}(\mX) + \log (-\mathcal{D}(\hat{\mX})) + \mathcal{L}_{rec}(\hat{\mX}),
\end{aligned}
\end{align}
where $\hat{\mX}$ is the generated image, $\mathcal{L}_{rec} = \sum_{l=1}^{L}\norm{\Phi^{(l)}(\hat{\mX})-\Phi^{(l)}(\mX))}_1$, $\Phi^{(l)}$ is the feature representation of a pre-trained network at scale $l\in\{1,\cdots,L\}$ whereas $L=4$. We use a pre-trained segmentation model with a large receptive field to compute features for large-mask inpainting~\cite{lama}. In order to generate the object bounding box for training the object-level discriminators, we extract maximal and minimal coordinates of a binary instance map on the fly during training.

\subsection{A Fully Automatic Pipeline for Standard Inpainting}
\label{sec:automantic_inpainting}
\revision{Guided inpainting approaches~\cite{deepfillv2,sesame,sketchedit} usually assume the presence of a guidance map. This reliance often complicates the direct application of guided inpainting methods to conventional image inpainting tasks. In this section, we introduce a novel and fully automatic pipeline that adapts our pre-trained guided inpainting model to standard inpainting scenarios.} Our pipeline to predict panoptic segmentation of the whole image given the incomplete image as input. More specifically, our process begins with utilizing an off-the-shelf inpainting method, such as CM-GAN~\cite{cmgan} to produce an initial completed image. We then apply PanopticFCN~\cite{panopticFCN} to this image to generate a completed panoptic layout. Subsequently, both the predicted panoptic layout and the masked image are fed into our guided inpainting model for further inpainting. As an alternative strategy, we demonstrate that it's feasible to train a specialized PanopticFCN model that directly predicts the panoptic segmentation of the whole image from an incomplete one. For additional details, please consult the supplementary material.

	\begin{table*}[]
		\small
		\caption{
			Evaluation of instance-guided inpainting, segmentation-guided inpainting, and edge-guided inpainting on Places2-person. Methods with $*$ are re-trained on the instance-guided inpainting tasks.
		}
		
		\centering
		\begin{tabular}{l|c|c|c|c|c|c}
			\toprule
			& \multicolumn{3}{c|}{CoModGAN masks} & \multicolumn{3}{c}{Object masks}\\
			\hline
			Methods & FID$\downarrow$ & U-IDS (\%)$\uparrow$ & P-IDS (\%)$\uparrow$ & FID$\downarrow$ & U-IDS (\%)$\uparrow$ & P-IDS (\%)$\uparrow$\\
			\hline
			\multicolumn{7}{l}{\B \em inpainting w/ instance segmentation}\\
			\hline
			SESAME$*$~\cite{sesame} & 12.0061 & 7.41 & 0.24 & 8.3656 & 9.91 & 0.41 \\
			LaMa$*$~\cite{lama} & 7.0563 & 21.74 & 4.83 & 4.8156  & 26.28 & 7.39 \\
			CoModGAN$*$~\cite{comodgan} & 5.0168 & 29.58 & 14.94 & 4.4232 & 31.59 & 16.75 \\
			CM-GAN$*$~\cite{cmgan} & 2.8470 & 33.20 & 18.50 & 2.7246  & 34.42 & 19.66 \\
			\revision{ControlNet$*$~\cite{controlnet}+RePaint~\cite{repaint}} & 7.4024 & 24.03 & 8.26 & 4.3714 & 28.82 & 11.56 \\
			\B ours  & \B 2.0720 & \B 36.96 & \B 25.90 & \B 1.8682 & \B 37.90 & \B 26.30  \\
			\hline
			\multicolumn{7}{l}{\B \em inpainting w/ segmentation}\\
			\hline
			SESAME~\cite{sesame} & 12.2308 & 7.09 & 0.22 & 8.3940 & 9.75 & 0.40 \\
			\revision{ControlNet$*$~\cite{controlnet}+RePaint~\cite{repaint}} & 7.4306 & 24.33 & 8.42 & 4.3910 & 29.13 & 12.13 \\
			\B ours & \B 2.3860 & \B 33.11 & \B 19.25 & \B 2.1565 & \B 34.85 & \B 21.12 \\
			\hline
			\multicolumn{7}{l}{\B \em inpainting w/ instance edge}\\
			\hline
			EdgeConnect~\cite{edgeconnect} & 41.7631 & 3.18 & 0.04 & 22.9517 & 3.98 & 0.06 \\
			SketchEdit~\cite{sketchedit} & 16.1271 & 13.02 & 1.58 & 8.7878 & 19.77 & 3.21 \\
			\revision{ControlNet$*$~\cite{controlnet}+RePaint~\cite{repaint}} & 8.5324 & 22.42 & 6.82 & 4.8271 & 27.48 & 10.19 \\
			\B ours & \B 2.6909 & \B 33.43 & \B 20.45 & \B 2.1873 & \B 36.19 & \B 23.46 \\
			\hline
			\multicolumn{7}{l}{\revision{\B \em inpainting w/ canny edge}}\\
			\hline
			\revision{EdgeConnect}~\cite{edgeconnect} & 47.9578 & 3.71 & 0.10 & 42.5151 & 2.67 & 0.04 \\
			\revision{SketchEdit}~\cite{sketchedit} & 14.5038 & 13.25 & 1.46 & 7.0494 & 20.60 & 3.03 \\
			\revision{ControlNet$*$~\cite{controlnet}+RePaint~\cite{repaint}} & 9.9058 & 20.30 & 4.90 & 5.3867 & 26.43 & 8.94 \\
			\B \revision{ours} & \B 1.6736 & \B 37.82 & \B 25.78 & \B 1.3116 & \B 40.11 & \B 28.71 \\
			\bottomrule
		\end{tabular}
		\label{tab:revision:main}
	\end{table*}

	\begin{table*}[]
	\small
	\caption{
		\revision{Evaluation of instance-guided inpainting, segmentation-guided inpainting, and edge-guided inpainting on the Coco-Stuff dataset. Methods with $*$ are re-trained on the instance-guided inpainting tasks.}
	}
	\centering
	\begin{tabular}{l|c|c|c|c|c|c}
		\toprule
		& \multicolumn{3}{c|}{CoModGAN masks} & \multicolumn{3}{c}{Object masks}\\
		\hline
		Methods & FID$\downarrow$ & U-IDS (\%)$\uparrow$ & P-IDS (\%)$\uparrow$ & FID$\downarrow$ & U-IDS (\%)$\uparrow$ & P-IDS (\%)$\uparrow$\\
		\hline
		\multicolumn{7}{l}{\B \em inpainting w/ instance segmentation}\\
		\hline
		\revision{SESAME$*$~\cite{sesame}} & 19.7117 & 3.08 & 0.26 & 13.7626 & 6.90 & 0.88 \\
		\revision{LaMa$*$~\cite{lama}} & 15.2794 & 18.59 & 6.23 & 10.1223 & 25.81 & 10.16 \\
		\revision{CoModGAN$*$~\cite{comodgan}} & 13.9974 & 25.73 & 14.37 & 10.2675 & 30.66 & 18.38 \\
		\revision{CM-GAN$*$~\cite{cmgan}} & 11.1592 & 28.49 & 17.36 & 9.1851  & 32.51 & 21.68 \\
		\revision{ControlNet$*$~\cite{controlnet}+RePaint~\cite{repaint}} & 16.6987 & 20.50 & 7.99 & 9.8972  &  28.68 & 15.60  \\
		\revision{\B ours}  & \B 10.1081 & \B 30.60 & \B 20.29 & \B 8.2422 & \B 33.84 & \B 23.68  \\
		\hline
		\multicolumn{7}{l}{\B \em inpainting w/ segmentation}\\
		\hline
		\revision{SESAME~\cite{sesame}} & 20.0830 & 2.26 & 0.12 & 13.7859 & 6.38 & 0.44 \\
		\revision{ControlNet$*$~\cite{controlnet}+RePaint~\cite{repaint}} & 20.4508 & 18.46 & 6.41 & \B 9.9252 & 28.78 & 15.95 \\
		\B \revision{ours}  & \B 13.6208 & \B 21.21 & \B 9.18 & 10.1776 & \B 27.55 & \B 15.60 \\
		\hline
		\multicolumn{7}{l}{\B \em inpainting w/ instance edge}\\
		\hline
		\revision{EdgeConnect~\cite{edgeconnect}} & 54.6826 & 0.02 & 0.00 & 29.1238 & 0.67 & 0.04 \\
		\revision{SketchEdit~\cite{sketchedit}} & 17.4214 & 15.54 & 4.67 & 10.3242 & 23.40 & 8.55\\
		\revision{ControlNet$*$~\cite{controlnet}+RePaint~\cite{repaint}} & 17.7643 & 19.14 & 7.07 & \B 10.2585 & 27.96 & 14.64 \\
		\B \revision{ours} & \B 16.5343 & \B 22.51 & \B 10.22 & 11.5286 & \B 28.36 & \B 16.12 \\
		\hline
		\multicolumn{7}{l}{\revision{\B \em inpainting w/ canny edge}}\\
		\hline
		\revision{EdgeConnect~\cite{edgeconnect}} & 59.7536 & 0.01 & 0.00 & 33.9190 & 0.03 & 0.00 \\
		\revision{SketchEdit~\cite{edgeconnect}} & 24.3690 & 10.32 & 2.08 & 31.9886 & 0.11 & 0.00 \\
		\revision{ControlNet$*$~\cite{controlnet}+RePaint~\cite{repaint}} & 18.3780 & 17.12 & 4.83 & \B 10.2005 & 26.71 & 11.82 \\
		\B \revision{ours} & \B 15.4551 & \B 22.84 & \B 10.48 & 10.6509 & \B 28.89 & \B 16.21 \\
		\bottomrule
	\end{tabular}
	\label{tabs:main-coco}
\end{table*}


\begin{table}[]
    \small
	\caption{
		Quantitative evaluation of instance-guided inpainting on Places2-object.
	}
	\centering
		\begin{tabular}{l|c|c|c}
			\toprule
			Methods & FID$\downarrow$ & U-IDS (\%)$\uparrow$ & P-IDS (\%)$\uparrow$\\
			\hline
            SESAME$*$~\cite{sesame} & 7.6420 & 11.92 & 0.64 \\
			LaMa$*$~\cite{lama} & 4.1189 & 31.05 & 11.35 \\
			CoModGAN$*$~\cite{comodgan} & 5.9140 & 31.39 & 15.44 \\
			CM-GAN$*$~\cite{cmgan} & 3.3929 & 36.02 & 20.92 \\
			\B ours & \B 3.2126 & \B 37.58 & \B 25.80 \\
			\bottomrule
		\end{tabular}
	\label{tab:main_general_object}
\end{table}


\begin{figure*}[t]
	\centering
	\includegraphics[width=1\linewidth]{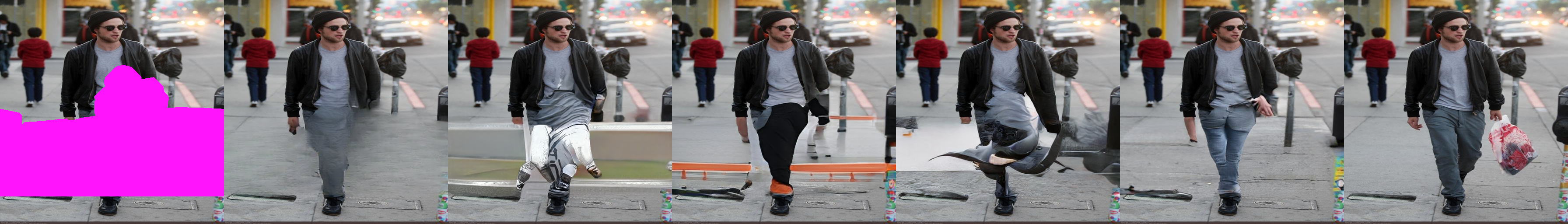}\\
	\includegraphics[width=1\linewidth]{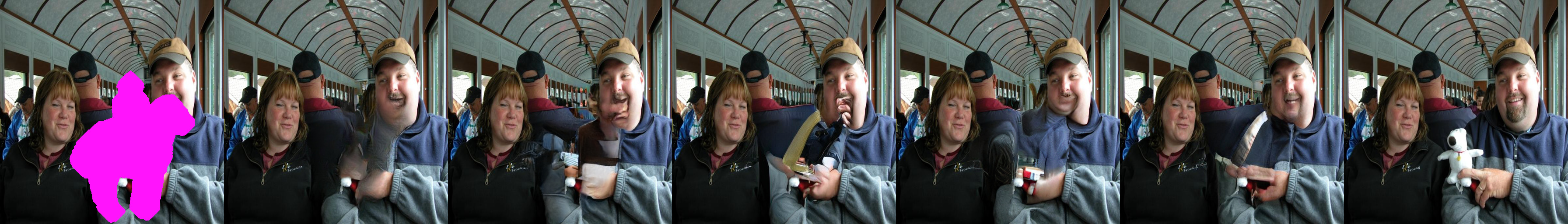}\\
	\includegraphics[width=1\linewidth]{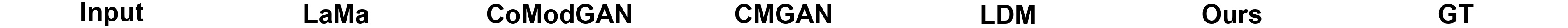}
	\caption{
		Qualitative comparisons on the standard inpainting task. Compared to the existing methods, our method can generate high-quality and photo-realist object instances.
	}
	\label{fig:pure_inpainting}
\end{figure*}

\section{Experiments}
\label{sec:experiment}
\subsection{Implementation Details}
\noindent \textbf{Datasets and evluation.} \quad
We collect two large-scale object-centric datasets named Places2-person and Places2-object from the Places2 dataset~\cite{zhou2017places} for evaluating various object inpainting tasks in various settings.
Specifically, Places2-person and Places2-object are subsets of the Places2 dataset that contain at least one person or general object instances, respectively.
\revision{Additionally, our model was trained and evaluated using the COCO-Stuff dataset~\cite{coco}, which comprises real-world images featuring numerous object instances and semantic annotations.
We leverage pre-trained PanopticFCN model~\cite{panopticFCN} to generate instance and semantic segmentation annotations for all the datasets and apply the random stroke mask~\cite{comodgan} and object-shaped masks~\cite{profill} for model evaluation. 
It should be noted that although the COCO-Stuff dataset includes accurate instance segmentation ground truth, we opted to use the PanopticFCN for generating annotations to bolster the generalization ability of our model in real-world image editing applications.
Furthermore, the edge-guided inpainting task is evaluated in two settings. In the first setting, the edge map is extracted by computing the boundary for the instance segmentation mask. In the second setting, the edge map is extracted by applying a pre-trained edge detection model~\cite{bdcn} followed by a canny edge detection step. The resulting two edge formats are denoted by \textit{instance edge} and \textit{Canny edge}, respectively.
}
We report the numerical metrics on test sets using the masking scheme of CoModGAN \cite{comodgan} and the object masks of~\cite{profill} and report \emph{Frchet Inception Distance} (FID)~\cite{fid} and the \emph{Paired/Unpaired Inception Discriminative Score} (P-IDS/U-IDS)~\cite{comodgan} for evaluation.

\noindent \textbf{Inpainting Tasks and compared methods.} \quad
We evaluate our model on the \emph{instance-guided}, \emph{segmentation-guided}, and \emph{edge-guided} inpainting tasks. Furthermore, \cref{sec:pure_inpainting} evaluates our fully automatic pipeline for the \emph{standard inpainting} task. For the instance-guided task, we compare our method with the recent inpainting and guided-inpainting methods including SESAME$*$~\cite{sesame}, LaMa$*$~\cite{lama}, CoModGAN$*$~\cite{comodgan} and CM-GAN$*$~\cite{cmgan}, where $*$ symbol denotes models retrained for the instance-guided task. 
All the retrained models are trained on 8 A100 GPUs for at least three days and until convergence to ensure fair comparisons.
For segmentation-guided inpainting, we compare our method with SESAME~\cite{sesame}, and for the edge-guided inpainting, we compare our method with Edge-connect~\cite{edgeconnect} and SketchEdit~\cite{sketchedit}.
\revision{In addition, we compare our approach with the recent ControlNet~\cite{controlnet} for guided inpainting and the Latent Diffusion inpainting model~\cite{LDM}. To tailor ControlNet for the guided inpainting task, we retrain ControlNet models on multiple guidance inputs, such as instance segmentation, semantic segmentation, and edge map for guided generation, then utilize RePaint~\cite{repaint} to paste back the known region, resulting in ControlNet models for guided inpainting. For ControlNet and Latent Diffusion, we set the text prompt to ``a clean, beautiful, and high-resolution image'' to avoid potential performance degradation.}


\begin{table*}[]
	\caption{
		Ablation studies of our model. \emph{Adv.}, \emph{perc.}, \emph{sem. D}, \emph{obj. D} are abbreviations of adversarial loss, perceptual loss, semantic discriminator and object-level discriminator, respectively.
	}
	\small
	\centering
		\begin{tabular}{l|c|c|c|c|c|c}
			\toprule
			& \multicolumn{3}{c|}{CoModGAN masks} & \multicolumn{3}{c}{Object masks}\\
			\hline
			Methods & FID$\downarrow$ & U-IDS (\%)$\uparrow$ & P-IDS (\%)$\uparrow$ & FID$\downarrow$ & U-IDS (\%)$\uparrow$ & P-IDS (\%)$\uparrow$\\
			\hline
			ours \textit{w/} adv. & 10.5587 & 20.60 & 5.56 & 9.3800 & 22.70 & 6.38\\
			ours \textit{w/} adv. + perc. & 2.8470 & 33.20 & 18.50 & 2.7246 & 34.42 & 19.66 \\
			\revision{ours \textit{w/} adv. + perc. + obj. D} & 2.8100 & 32.80 & 18.58 & 2.3477 & 35.56 & 21.74\\
			ours \textit{w/} adv. + perc. + sem. D & 2.2705 & 35.41 & 22.30 & 2.1636 & 36.12 & 22.98 \\
			\B ours \textit{w/} adv. + perc. + sem. D + obj. D (full) & \B 2.0720 & \B 36.96 & \B 25.90 & \B 1.8682 & \B 37.90 & \B 26.30 \\
			ours full \textit{w/} semantic segmentation & 2.3860 & 33.11 & 19.25 & 2.1565 & 34.85 & 21.12 \\
			\bottomrule
		\end{tabular}
	\label{tab:ablation}
\end{table*}


\begin{figure*}[h]
	\centering
	\includegraphics[width=1.\linewidth]{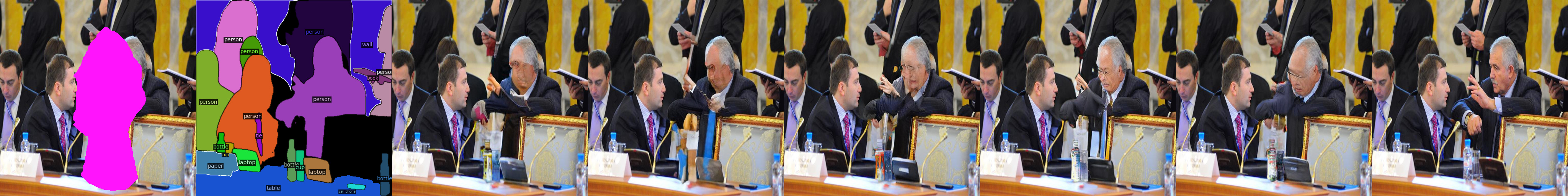}
	\includegraphics[width=1.\linewidth]{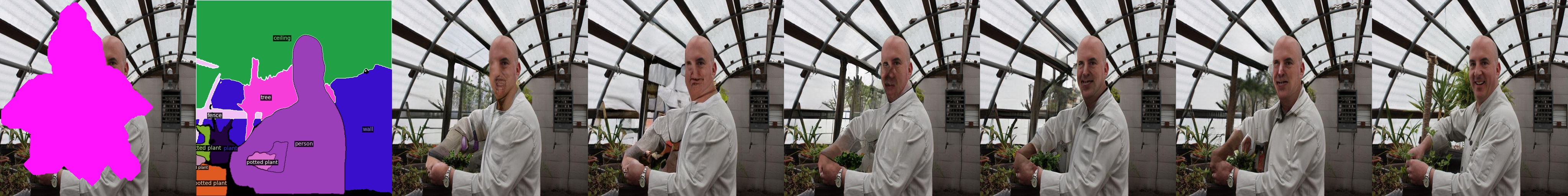}
	\includegraphics[width=1.\linewidth]{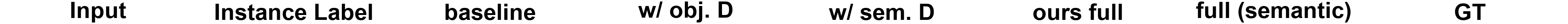}
	\caption{
	\small{
		\revision{he visual effect of various ablation models. Compared to the baseline model trained with StyleGAN discriminator and perceptual loss, our object-discriminator (obj. D), semantic discriminator (sem. D), and the combination of both progressively improve the realism of the generated objects and achieve the best visual results. Best viewed by zoom-in on screen.}
	}
	}
	\label{fig:ablation}
\end{figure*}

\noindent \textbf{Network Details.} \quad
We leverage the pretrained CLIP model~\cite{clip} as a backbone to extract features for semantic discriminators. Please refer to the appendix for more details about the network structure and training configuration.




\subsection{Quantitative and Qualitative Evaluation}
\Cref{tab:main_inpainting} and \Cref{tabs:main-coco} presents the evaluation on the instance-, segmentation-, and edge-guided inpainting tasks on Places2-person and Coco-Stuff datasets, respectively.
For all the tasks, our method achieves significant gains compared to the existing methods.
In addition, we observe that our instance-guided model achieves better FID scores compared to segmentation-guided or edge-guided counterparts thanks to the instance-level semantic information provided by the panoptic guidance. However, our segmentation-guided and edge-guided models still achieve impressive FID scores compared to the existing methods, showing the flexibility and robustness of our approach.
Furthermore, \cref{tab:main_general_object} presents the evaluation of the instance-guided task on Places2-object where our model improves the existing methods and shows generalization capacity to general object classes. \revision{In addition, we observe that the guided inpainting with canny edge maps achieves better performance than its counterparts with instance edge maps, as canny edge maps provide more structural details for guided inpainting.}

\begin{figure*}[t]
	\centering
	\small
	\includegraphics[width=1.6\columnwidth]{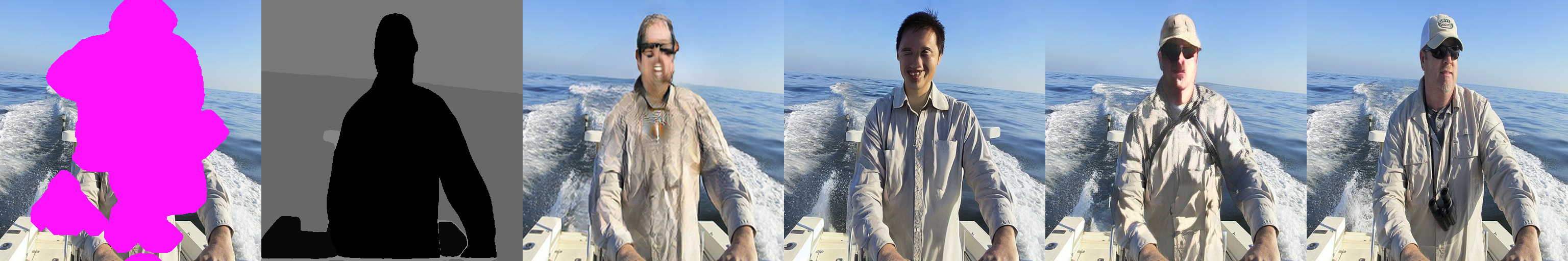}\\
	\includegraphics[width=1.6\columnwidth]{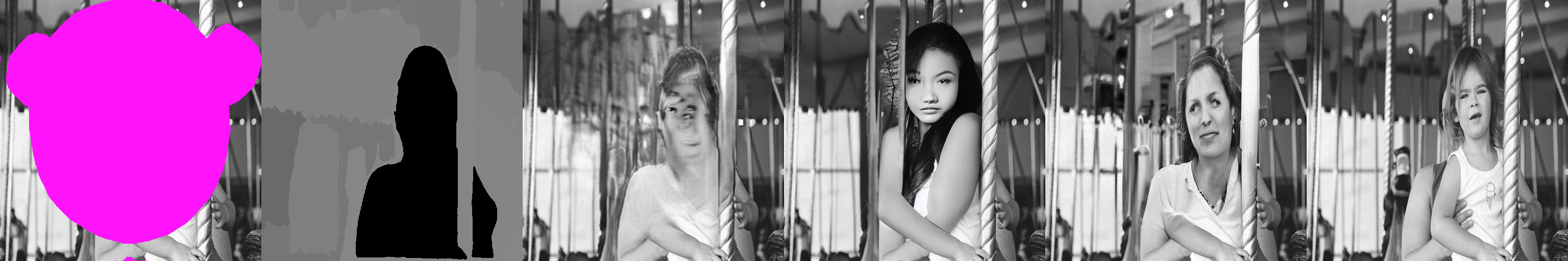}\\
	\vspace{-1mm}
	\includegraphics[width=1.6\columnwidth]{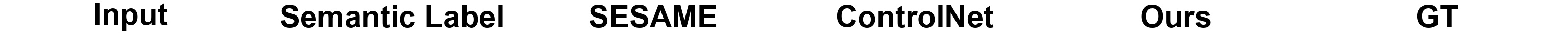}\\
	\includegraphics[width=1.6\columnwidth]{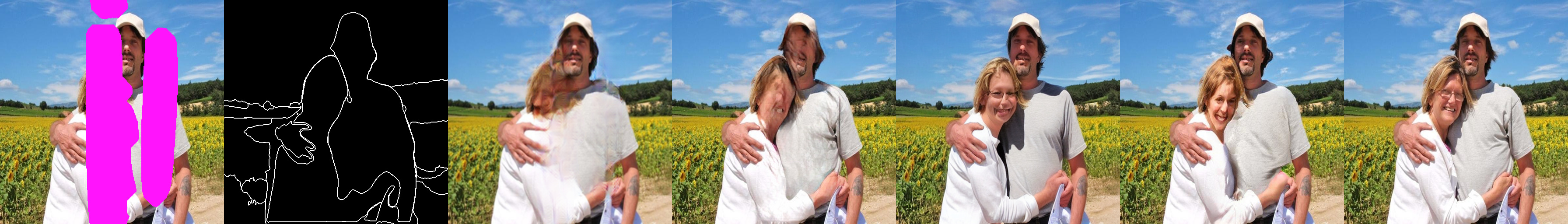}\\
	\includegraphics[width=1.6\columnwidth]{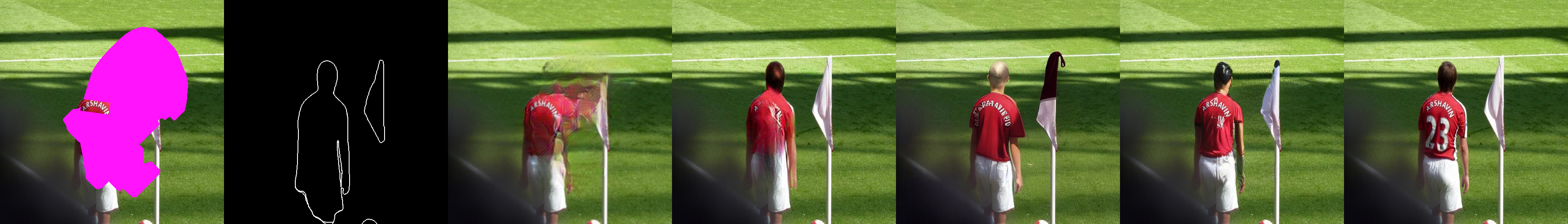}\\
	\vspace{-1mm}
	\includegraphics[width=1.6\columnwidth]{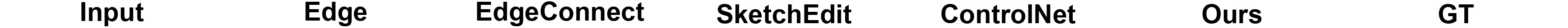}\\
	\includegraphics[width=1.6\columnwidth]{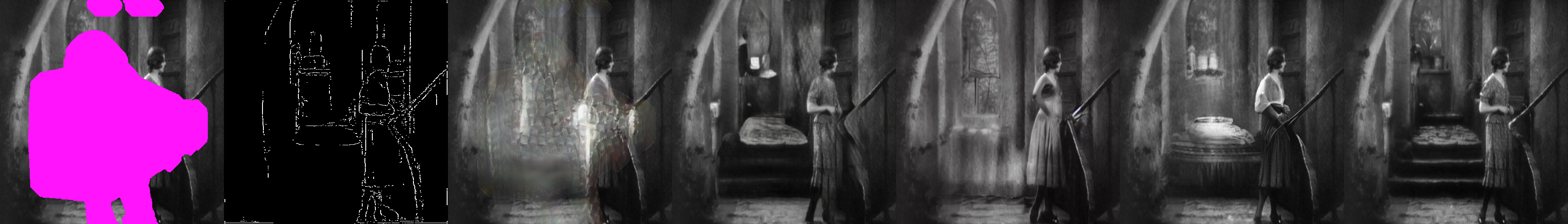}\\
	\includegraphics[width=1.6\columnwidth]{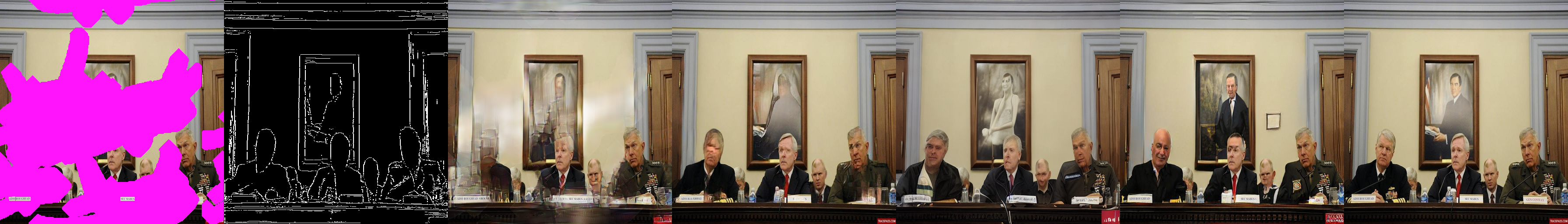}\\
	\vspace{-1mm}
	\includegraphics[width=1.6\columnwidth]{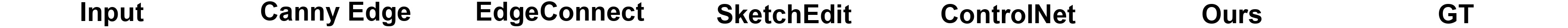}%
	\caption{
		\revision{Qualitative comparisons on segmentation-guided inpainting (top) and instance edge-guided inpainting (middle) and canny edge-guided inpainting (bottom).}
	}
	\label{fig:semantic_edge_guided_inpainting}
\end{figure*}

To understand the visual effect of our approach, we present visual comparisons of our method with state-of-the-art methods on the guided tasks.
Specifically, \cref{fig:main_panoptic} presents the qualitative comparisons of our method on the panoptically guided inpainting task on Places2-person against the retrained SESAME~\cite{sesame}, LaMa~\cite{lama}, CoModGAN~\cite{comodgan} and CM-GAN~\cite{cmgan} and ControlNet where \Cref{fig:main_panoptic_object} shows the visual comparison on Places2-object.
The additional visual comparisons on segmentation-guided and edge-guided tasks on Places2 datasets are presented in \cref{fig:semantic_edge_guided_inpainting}.
Furthermore, the various guided-inpainting comparison on COCO-Stuff dataset is presented in \cref{fig:main_coco}.
The visual comparison 
demonstrate the clear advantage of our method for generating realistic object instances compared to the GAN-based approach, including recent works such as CM-GAN~\cite{cmgan}, SESAME~\cite{sesame} and SketchEdit~\cite{sketchedit}. Moreover, our GAN-based approach can match object generation quality compared to the recent diffusion-based approach ControlNet~\cite{controlnet}.

Furthermore, we conducted compared the inference-stage speed and model size of our model, compared to it's diffusion-base conterpart. Specifically, \cref{table:profile} details the inference-stage profiling for both our methods and ControlNet. Owing to the one-step inference process of our approach, it is markedly faster during the inference stage than the diffusion-based ControlNet, while still achieving similar or superior visual quality. Furthermore, the smaller size of our model offers additional benefits for deployment on resource-limited devices, such as mobile phones.

\begin{table}[]
\caption{
    Comparison of inference time between our method and the diffusion-based counterpart. We averaged the inference times over multiple runs, evaluating for 100 DDIM steps with a jump length of 5~\cite{RePaintScheduler}. The profiling was conducted on a single NVIDIA A100 GPU.
}
\centering
    \begin{tabular}{l|c|c|c}
        \toprule
        Methods & Model Size & Inference Times & Inference Steps\\
        \hline
            ControlNet~\cite{controlnet} & 361M & 3 min 25 sec & 500\\
            ~~~~~+RePaint~\cite{lugmayr2022repaint} &  &  & \\
        Ours & 75M & 0.279 sec & 1\\
        \bottomrule
    \end{tabular}
    \label{table:profile}
\end{table}

\subsection{Ablation Study}
We perform a set of ablation experiments to show the importance of each component of our model. Quantitative results are shown in \cref{tab:ablation} and the visual comparisons are shown in \cref{fig:ablation}. Below we describe the ablation experiments for the following components:

\noindent \textbf{Perceptual Loss} \quad
We start with the conditional CM-GAN as the baseline for the panoptically guided inpainting task. We find that the model trained with only the StyleGAN discriminator loss (abbreviated as adv.) suffers from slow convergence and sometimes produces color blobs, while the perceptual loss model (perc.) improves the performance and reduce the FID from 10.5587/9.3800 to 2.8470/2.2746, respectively, on the two masks. This finding is consistent with the observations of CM-GAN~\cite{cmgan} and LaMa~\cite{lama}.

\noindent \textbf{Semantic Discriminator} \quad
Based on the perceptual loss model, we insert the semantic discriminator (sem. D) at the image-level only for model training. As shown in \cref{tab:ablation}, the semantic discriminator improves the FID, which is consistent with the improvement in object generation, such as faces. However, the semantic discriminator model still suffers from object distortion.

\noindent \textbf{Object-level Discriminators} \quad
\revision{We also evaluated the impact of object-level discriminators (obj. D). We observed that integrating only the object discriminator into our base model leads to a marginal improvement in FID scores. However, a combination of the object and semantic discriminators yields the best performance. This suggests that additional semantic information may be essential for the object-level discriminator to effectively discern object structure. The visual results presented in \cref{fig:simple_ablation,fig:ablation} demonstrate the significant improvements in image quality brought about by the object-level and semantic discriminators.}

\noindent \textbf{Semantic Label Map Guidance} \quad
We compare the instance-guided task with the segmentation-guided task using our trained models. The instance guided task achieves better FID scores and generates better object boundaries when instances overlap (e.g. the two overlapping persons in \cref{fig:ablation}). Our segmentation-guided model produces high-quality results for disjoint instances.

\begin{table}[h]
	\caption{
		Quantitative evaluation on the standard inpainting task on Places2-person. Our fully automatic pipeline proposed in \cref{sec:automantic_inpainting}, i.e., \emph{ours (automatic)}, achieves better numerical results.
		}
	\small
	\centering
		\begin{tabular}{l|c|c|c}
			\toprule
			Methods & FID$\downarrow$ & U-IDS (\%)$\uparrow$ & P-IDS (\%)$\uparrow$ \\
			\hline
			\multicolumn{4}{l}{\B \em CoModGAN masks}\\
			\hline
			LaMa~\cite{lama} & 32.9607 & 8.08 & 0.66 \\
			CoModGAN~\cite{comodgan} & 12.0215 & 19.02 & 5.46\\
			CM-GAN~\cite{cmgan} & 11.6727 & 19.56 & 5.53 \\ 
			Latent Diffusion~\cite{ldm} & 12.8629 & 19.16 & 5.27\\ 
			\B ours (automatic) & \B 4.5402 & \B 29.65 & \B 22.42 \\ 	
			\hline
			\multicolumn{4}{l}{\B \em Object masks}\\
			\hline
			LaMa~\cite{lama} & 13.5481 & 15.91 & 2.46\\
			CoModGAN~\cite{comodgan} & 10.4286 & 20.59 & 5.38\\
			CM-GAN~\cite{cmgan} & 9.0216 & 23.05 & 15.18 \\ 
			Latent Diffusion~\cite{ldm} & 8.2006 & 24.14 & 8.17\\
			\B ours (automatic) & \B 3.1960 & \B 33.98 & \B 28.24 \\  
			\bottomrule
		\end{tabular}
	\label{tab:main_inpainting}
\end{table}

\subsection{Fully Automatic Image Inpainting} \label{sec:pure_inpainting}
\cref{tab:main_inpainting} presents the evaluation of our fully automatic inpainting pipeline (\cref{sec:automantic_inpainting}) in comparison to the recent state-of-the-art inpainting methods including LaMa~\cite{lama}, CoModGAN~\cite{comodgan}, CM-GAN~\cite{cmgan} and the recent latent diffusion inpainting model (Latent Diffusion)~\cite{ldm}.
Our approach significantly improves the numerical metrics with decreased FID scores, which is consistent with the improved realism of the generated objects as depicted in \cref{fig:pure_inpainting}.


\section{Conclusion}
Aiming at inpainting realistic objects, we investigate a structure-guided image inpainting task that leverages structure information such as semantic or panoptic segmentation to assist image inpainting. Our approach is based on a new semantic discriminator design that leverages pretrained visual features to improve the semantic consistency of the generated contents. We further propose  object-level discriminators to enhance the realism of the generated content. Our approach shows significant improvements on the generated objects and  leads to  new state-of-the-art performance on various tasks, including panoptic, semantic segmentation or edge map guided inpainting, and standard inpainting.


\appendices
\section{Dataset Details}
We construct two large-scale datasets named Places2-person and Places2-object from the Places2 dataset~\cite{zhou2017places} for evaluating the object inpainting task in various settings.
Specifically, Places2-person is the subsets of Places2 dataset that contains at least one person instances and it includes 1.28M images for training and 62748 images for testing. The Places2-object is the Places2 subset that contains at least one object instances and it includes 2.75M images for training and 127567 images for testing.
We leverage the pretrained PanopticFCN model \cite{panopticFCN} trained on COCO-Stuff \cite{coco} to generate the panoptic segmentation annotation on both datasets.
We use the inpainting mask of CoModGAN \cite{comodgan} to generate the mask for training and the mask of CoModGAN and the object mask of \cite{profill} for evaluation.
During training, we augment input images by random cropping and random horizontal flipping. 

\section{Architecture and Training Details}
We leverage the pretrained CLIP model~\cite{clip} as the backbone to extract feature for the semantic discriminators. During training, we randomly sample one object instance that overlaps the mask, then crop and resize object instances to resolution $224\times224$ for training the object-level StyleGAN and semantic discriminators. We use the panoptic segmentation annotation to generate the bounding box of objects. Specifically, for each instances, we take the minimal bounding box corresponding to the instance as the bounding box for cropping. To reduce aliasing and ringing artifacts of generated objects, we following the data augmentation practice of StyleGAN-ada \cite{stylegan_ada} and upscale the global image by a factor of  $2$ and then apply a band-limited low-pass filter before the cropping and resizing operations. To generate the cropped patch of discrete condition such as semantic map and edge map, we follow the same upscale-cropping-resizing routine that is used for generating object patches. However, we apply nearest sampling operation instead of filtering to produce discrete label map and edge map. Our codebase is implemented based on the Pytorch implementation of StyleGAN2-ada~\cite{stylegan_ada}. Our model is trained with the Adam optimizer~\cite{adam} with a learning rate of 0.001 and a batch size of 32. The model training takes 3.5 days to converge on a server with 8 A100 GPUs.

\section{Discussion on Automatic Inpainting Heuristics}
\label{sec:supp_discussion_automantic}
\begin{figure}[]
	\centering
	\includegraphics[width=1.\linewidth]{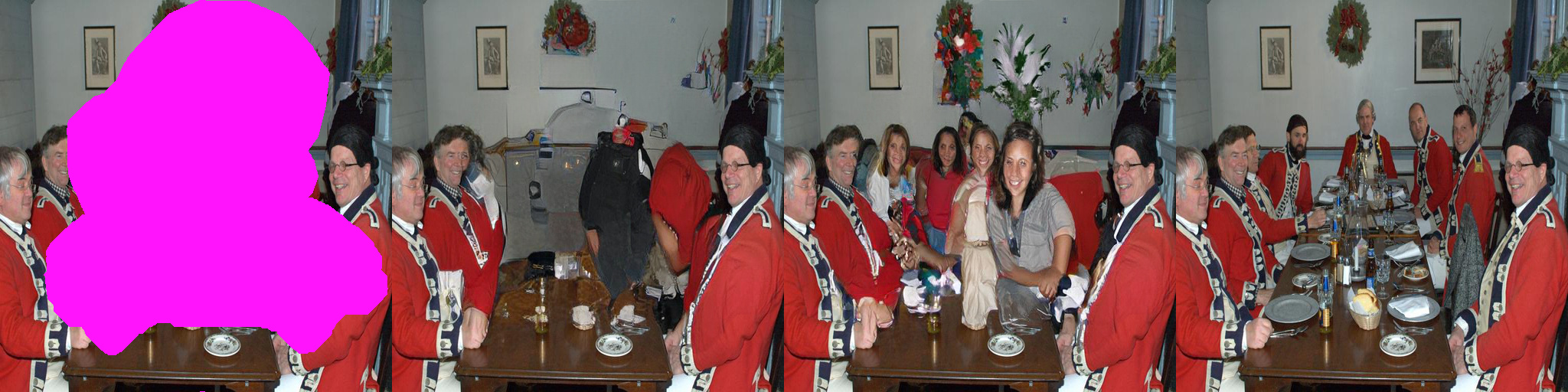}
	\includegraphics[width=1.\linewidth]{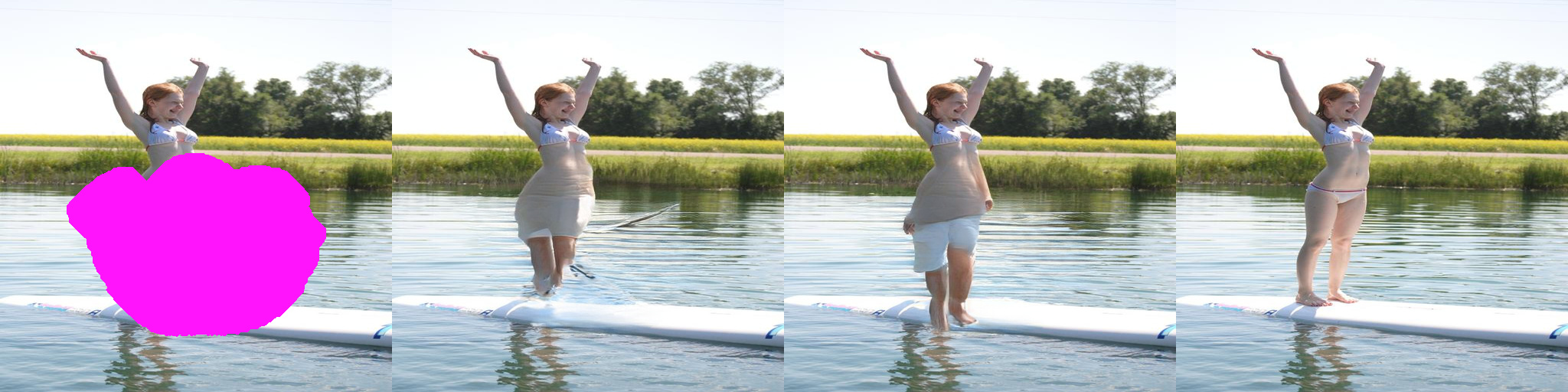}
	\includegraphics[width=1.\linewidth]{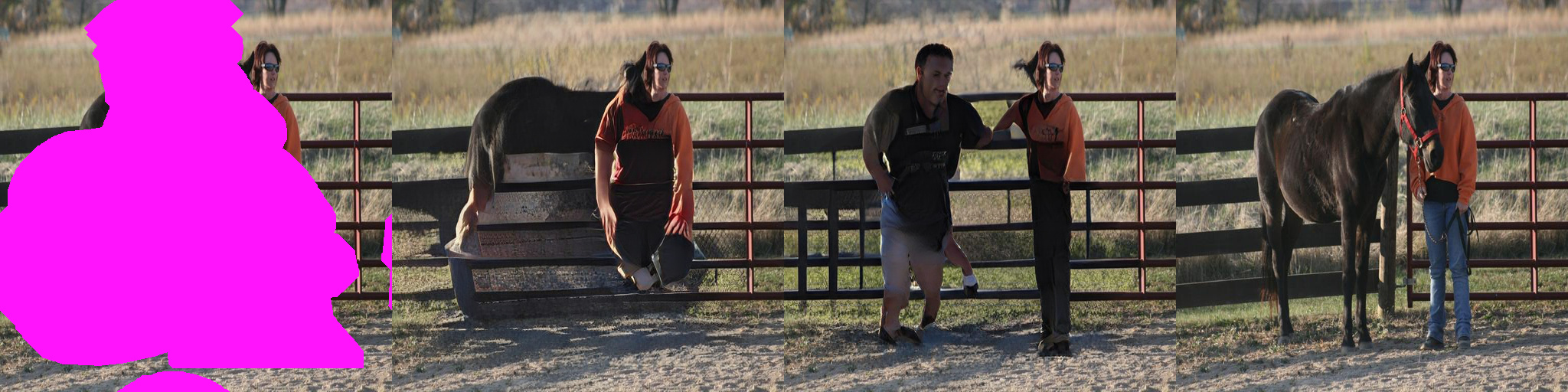}
	\includegraphics[width=1\linewidth]{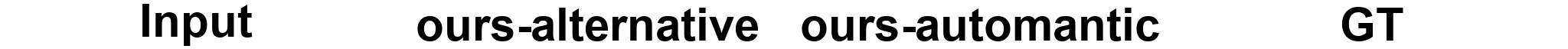}
	\caption{
		Visual comparisons of the different fully automatic inpainting pipeline. Our automatic pipeline \emph{ours (automantic)} proposed in the main paper generates more visually pleasing instances in comparison to its counterpart \emph{ours (alternative)} proposed in \cref{sec:supp_discussion_automantic} of the supplementary.
	}
	\label{fig:supp_pipeline_compare}
\end{figure}

In addition to the automatic inpainting pipeline that was proposed in Section 3.2, we propose an alternative inpainting pipeline. Specifically, we first train a PanopticFCN that takes the incomplete image as input and outputs the completed panoptic label map.
Next, the predicted panoptic layout and the masked image are passed to our guided inpainting model for automatic image inpainting. \cref{tab:supp_inpainting_pipeline} presents the numerical comparison of the propose alternative approach (denoted by \texttt{ours (alternative)}) in comparison to the approach proposed in the main paper (denoted by \texttt{ours (automantic)}).
In addition, \cref{fig:supp_pipeline_compare} presents the visual comparison of the two approaches. From the results, \texttt{ours (automantic)} tends to generate more realistic instance shape compared to \texttt{ours (alternative)} thanks to the generative modeling of the off-the-shelf inpainting model. Moreover, as amodal panoptic segmentation~\cite{mohan2022amodal} is a growing area,  our automatic inpainting pipeline could benefit from with stronger panoptic completion model.

\begin{table}[h]
	\caption{
		Quantitative comparison of different fully automatic pipelines for standard image inpainting.
		}
	\small
	\centering
		\begin{tabular}{l|c|c|c}
			\toprule
			Methods & FID$\downarrow$ & U-IDS (\%)$\uparrow$ & P-IDS (\%)$\uparrow$ \\
			\hline
			\multicolumn{4}{l}{\B \em CoModGAN masks}\\
			\hline
			ours (alternative) & 4.7807 & 29.18 & 21.94 \\ 	
			\B ours (automatic) & \B 4.5402 & \B 29.65 & \B 22.42 \\ 	%
			\hline
			\multicolumn{4}{l}{\B \em Object masks}\\
			\hline
			ours (alternative) & 6.8749 & 25.75 & 18.24 \\ 
			\B ours (automatic) & \B 3.1960 & \B 33.98 & \B 28.24 \\  
			\bottomrule
		\end{tabular}
	\label{tab:supp_inpainting_pipeline}
\end{table}

\section{Comparison to RePaint~\cite{lugmayr2022repaint}}
In addition, to supplement the evaluation from Table 4, we provide comparison to the recent diffusion-based inpainting method RePaint~\cite{lugmayr2022repaint} on the standard inpainting task. 
As the sampling procedure of RePaint is extremely slow, we report the performance on a subsets of places2-person that contains 4k samples.
Both the numerical results in \cref{tab:supp_main_inpainting} and visual results confirmed that our approach outperform RePaint  \cref{fig:supp_pure_inpainting}.

\section{More Visual Results}
\begin{figure*}[h]
	\centering
	\includegraphics[width=1.\linewidth]{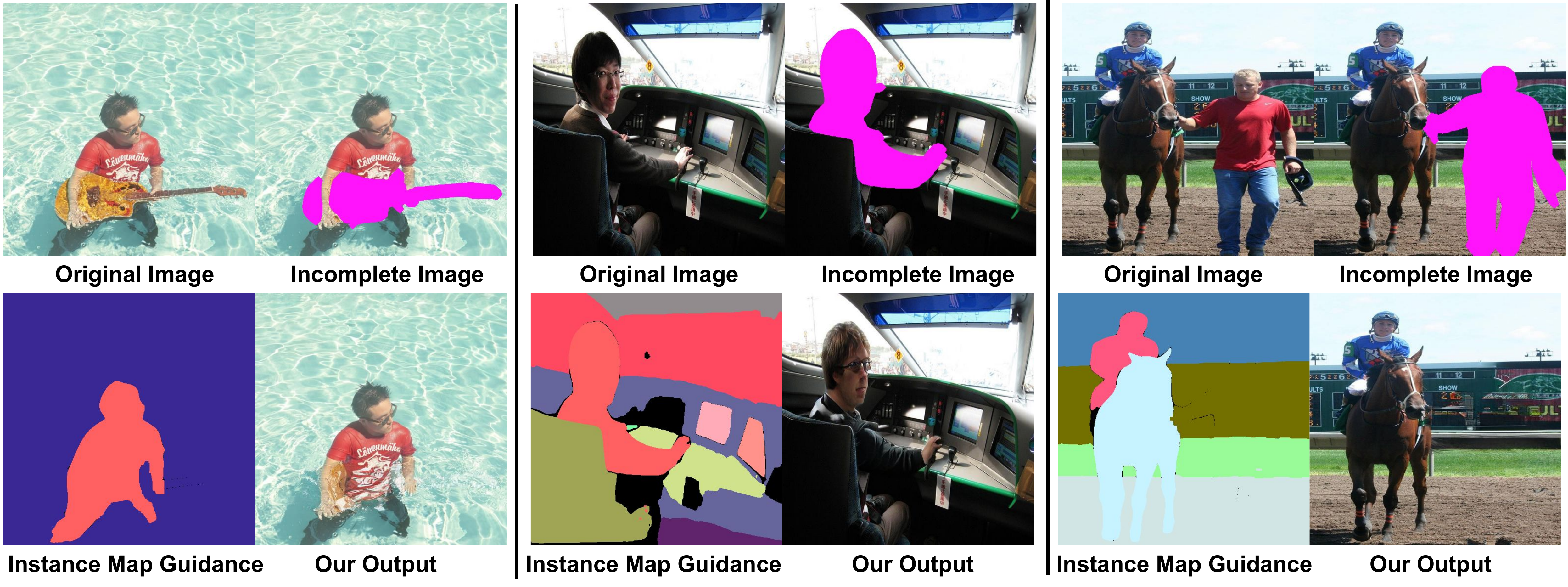}
	\caption{
		More image manipulation examples generated by our instance segmentation guided completion model. We showcase use cases such as local image manipulation (left), person anonymization (middle) and instance removal (right). Best viewed by zoom-in on screen.
	}
	\label{fig:supp_manipulation}
\end{figure*}

\begin{figure*}[h!]
	\centering
	\includegraphics[width=1.\linewidth]{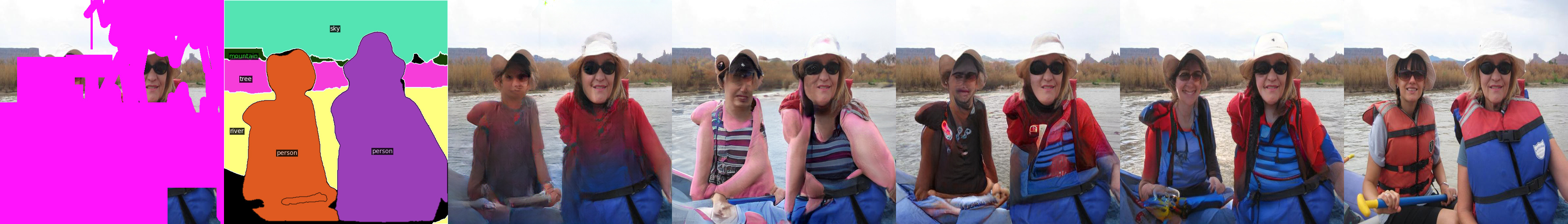}
	\includegraphics[width=1.\linewidth]{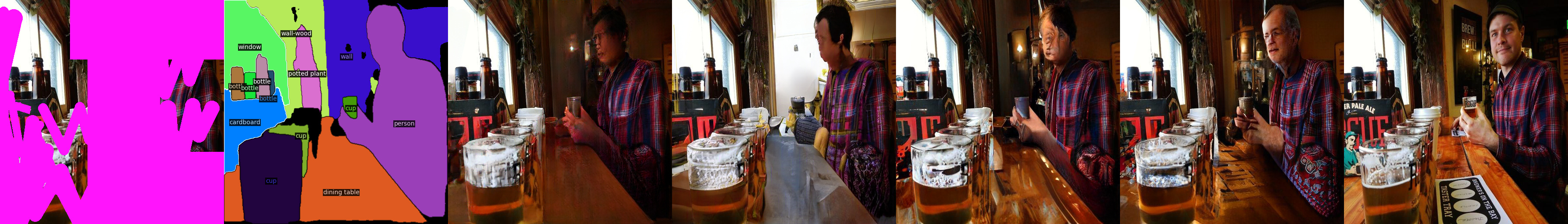}
	\includegraphics[width=1.\linewidth]{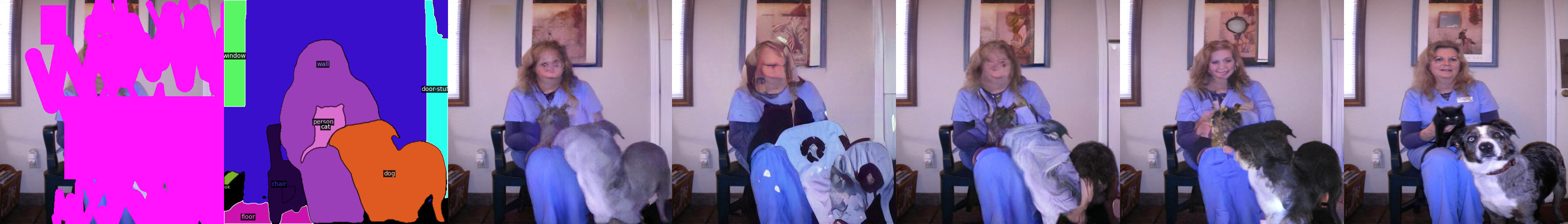}
	\includegraphics[width=1.\linewidth]{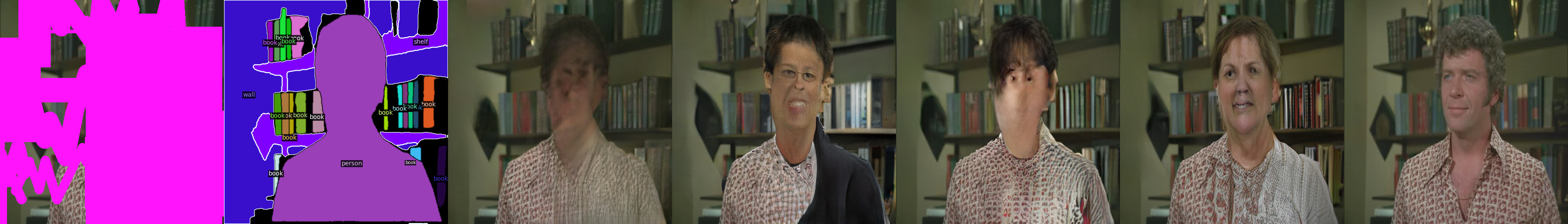}
	\includegraphics[width=1\linewidth]{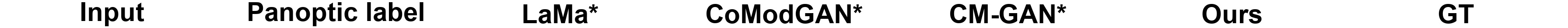}
	\caption{
		Visual comparisons on the instance segmentation guided inpainting task on Places2-person. We compare our model against LaMa*~\cite{lama}, CoModGAN*~\cite{comodgan}, CM-GAN*~\cite{cmgan} whereas  $*$  denotes models re-trained with the additional panoptic segmentation condition for instance segmentation guided inpainting. Best viewed by zoom-in on screen.
	}
	\label{fig:supp_main_panoptic}
\end{figure*}

\begin{figure*}[t]
	\centering
	\includegraphics[width=1\linewidth]{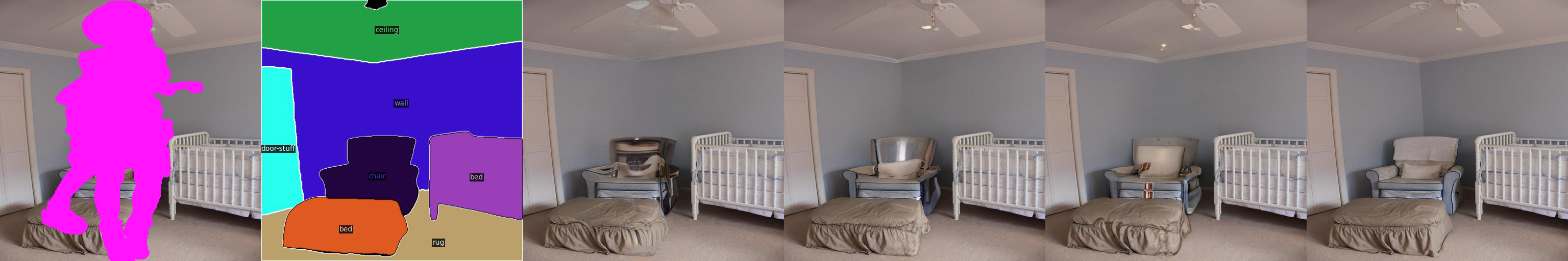}\\
	\includegraphics[width=1\linewidth]{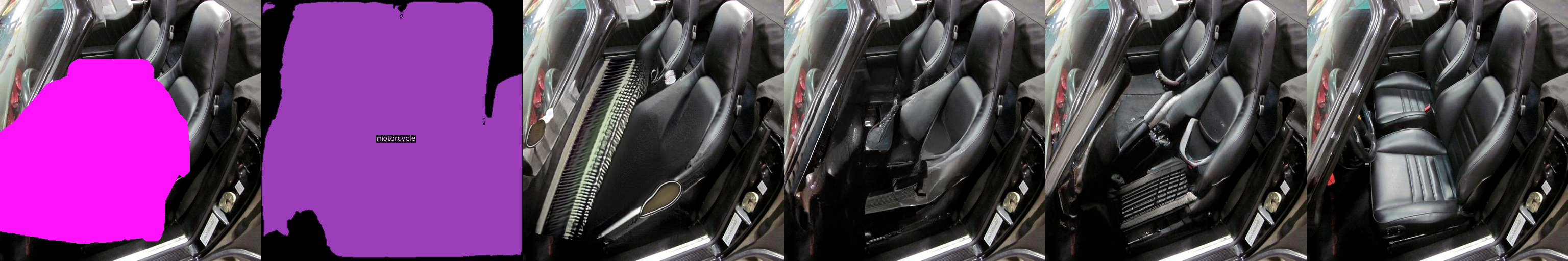}\\
	\includegraphics[width=1\linewidth]{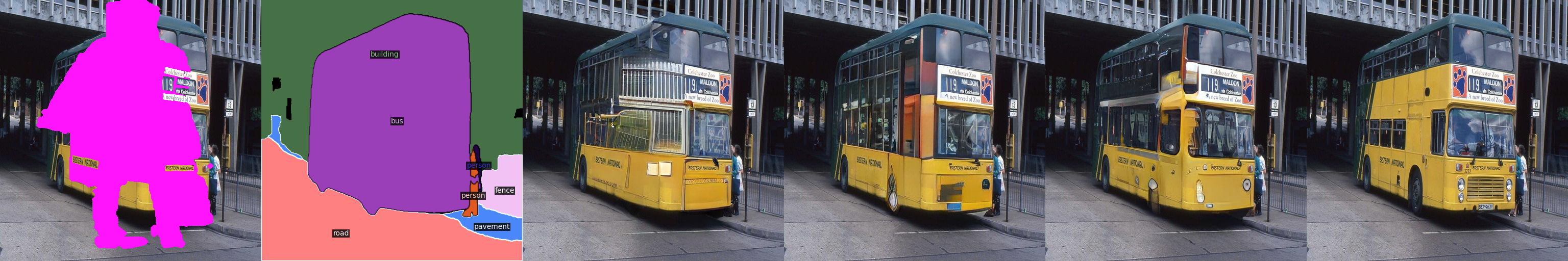}\\
	\includegraphics[width=1.0\linewidth]{figures_jpg/_vis_general_object_compare/_vis_general_object_compare_title_0.jpg}
	\caption{
		Qualitative comparisons on the instance guided inpainting task on Places2-object. We compare our model against retrained CoModGAN$*$~\cite{comodgan}, CM-GAN$*$~\cite{cmgan} whereas the $*$ symbol denotes models re-trained with the additional panoptic segmentation condition for instance guided inpainting. Best viewed by zoom-in on screen.
	}
	\label{fig:supp_main_panoptic_object}
\end{figure*}

\begin{figure*}[h!]
	\centering
	\includegraphics[width=1.\linewidth]{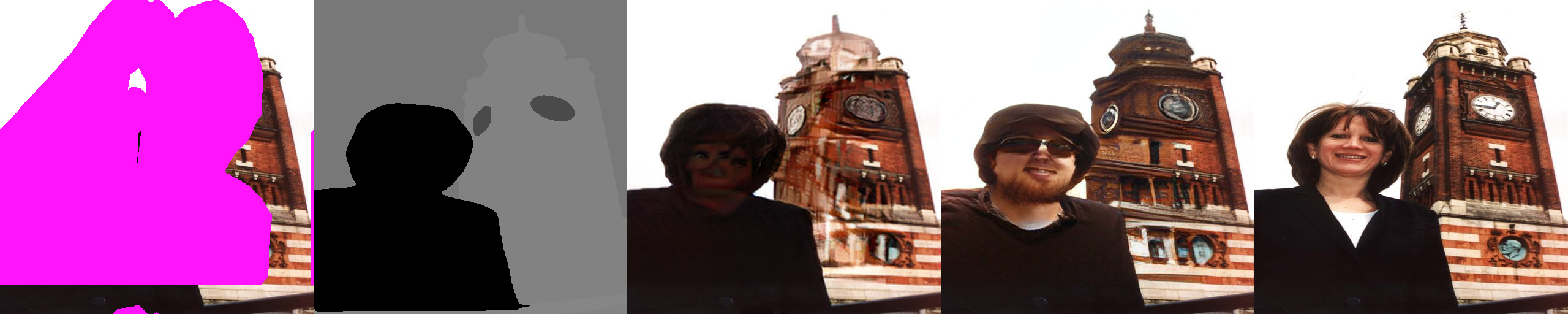}\\
	\includegraphics[width=1.\linewidth]{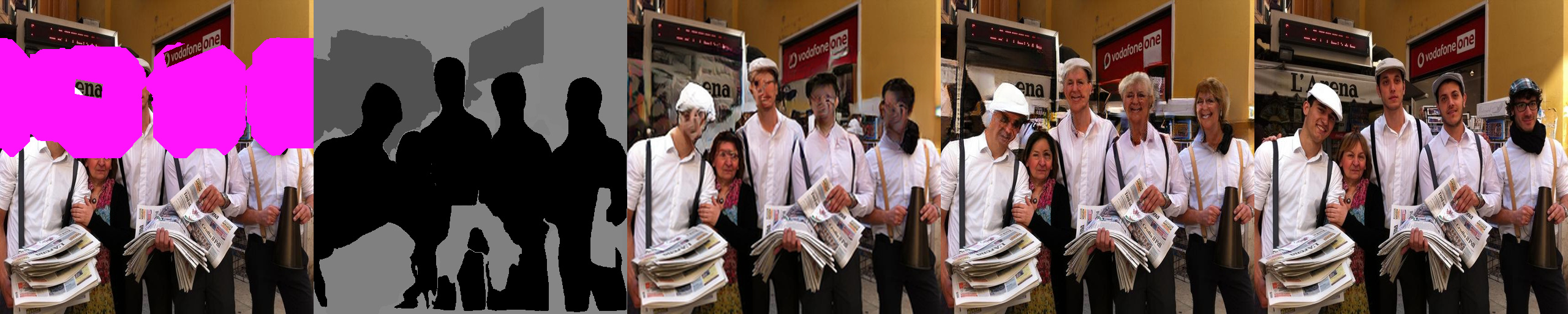}\\
	\includegraphics[width=1.\linewidth]{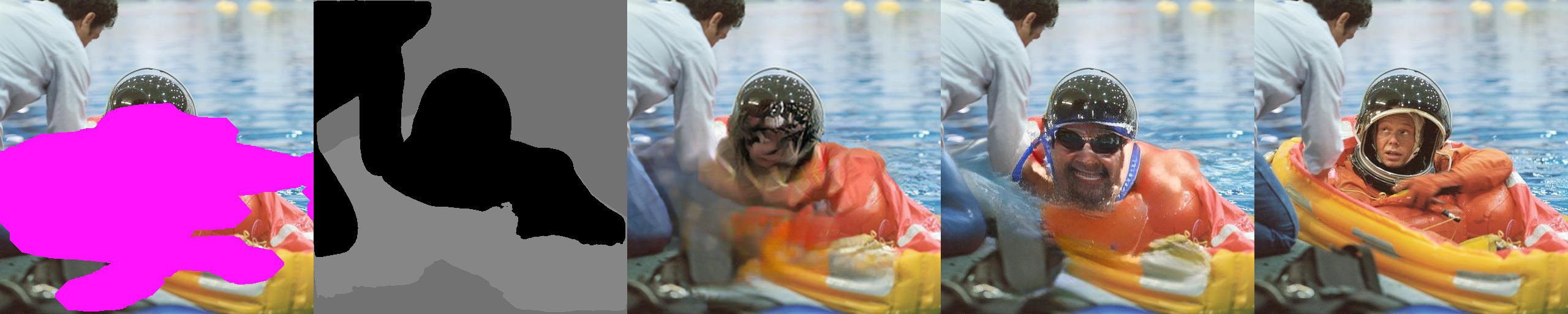}\\
	\includegraphics[width=1.\linewidth]{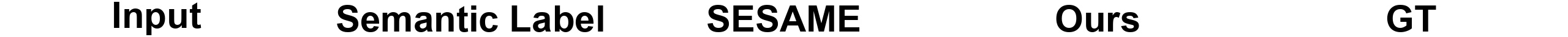}
	\caption{
		Qualitative comparisons on semantic-guided inpainting. Best viewed by zoom-in on screen.
	}
	\label{fig:supp_semantic_guided_inpainting}
\end{figure*}

\begin{figure*}[t]
	\centering
	\includegraphics[width=1.\linewidth]{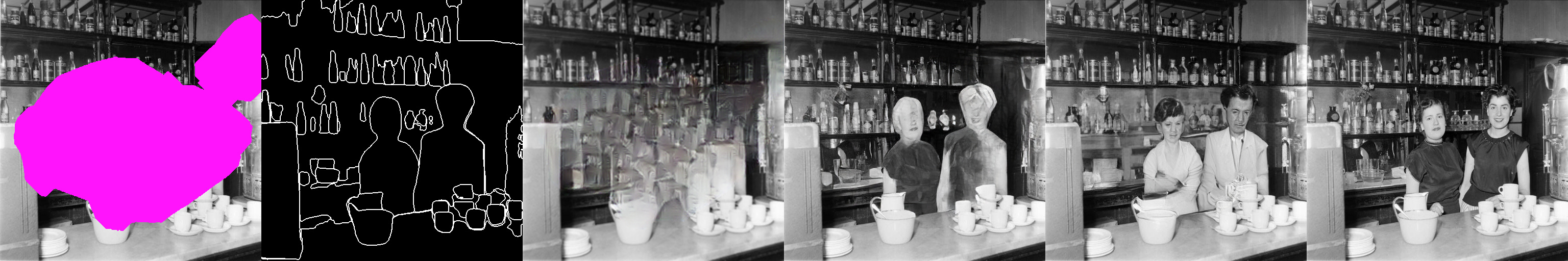}\\
	\includegraphics[width=1.\linewidth]{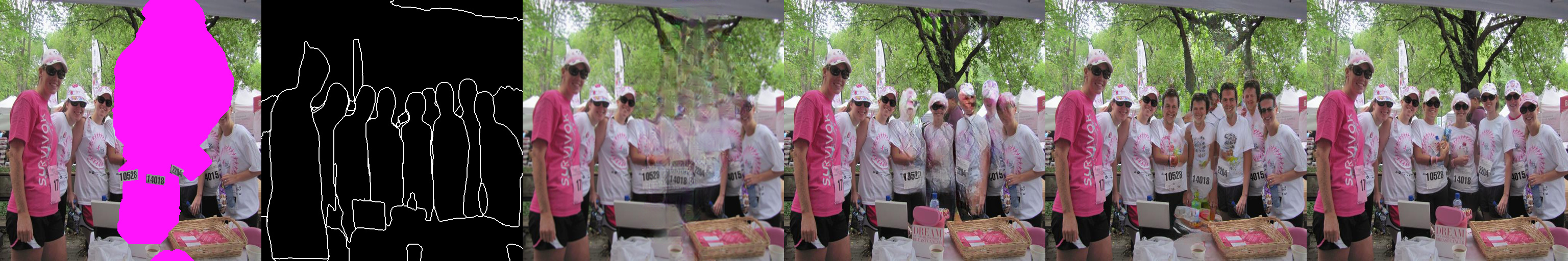}\\
	\includegraphics[width=1.\linewidth]{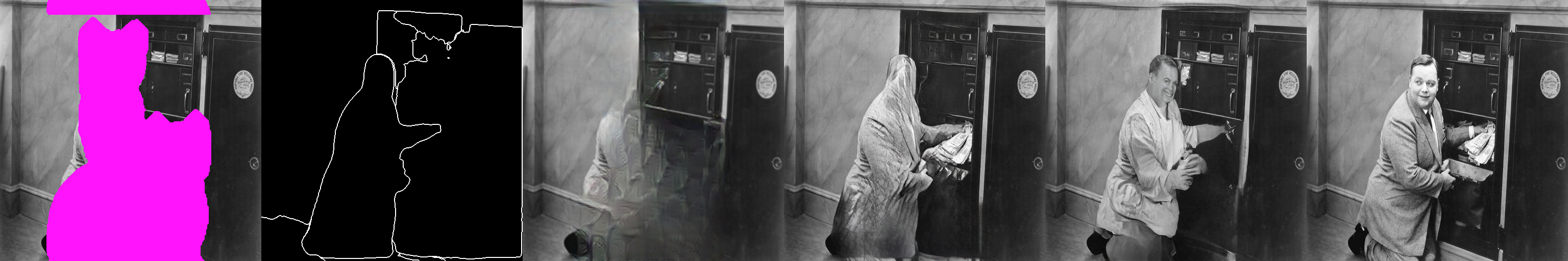}\\
	\includegraphics[width=1.\linewidth]{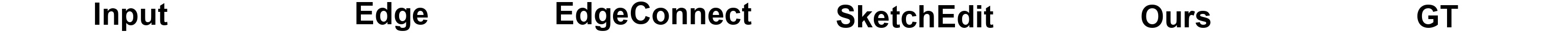}
	\caption{
		Qualitative comparisons on edge-guided inpainting.
		Best viewed by zoom-in on screen.
	}
	\label{fig:supp_edge_guided_inpainting}
\end{figure*}

\begin{figure*}[t]
	\centering
	\includegraphics[width=1.\linewidth]{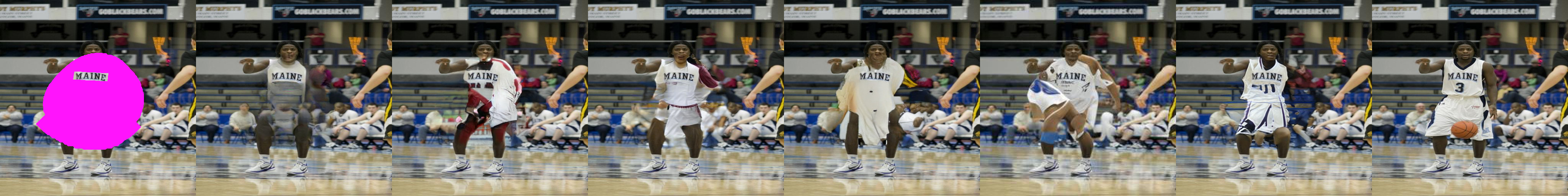}\\
	\includegraphics[width=1.\linewidth]{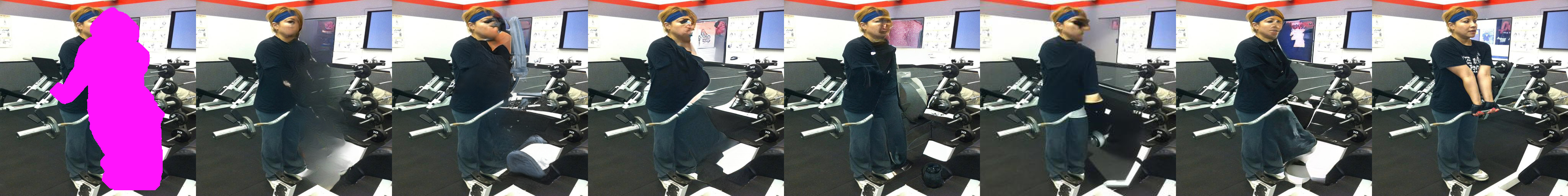}\\
	\includegraphics[width=1.\linewidth]{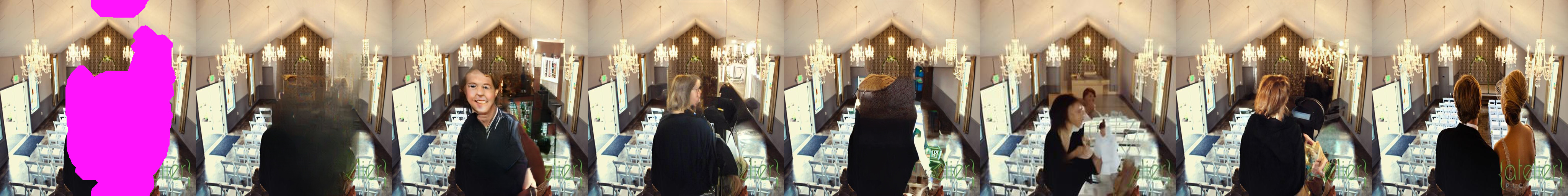}\\
	\includegraphics[width=1.\linewidth]{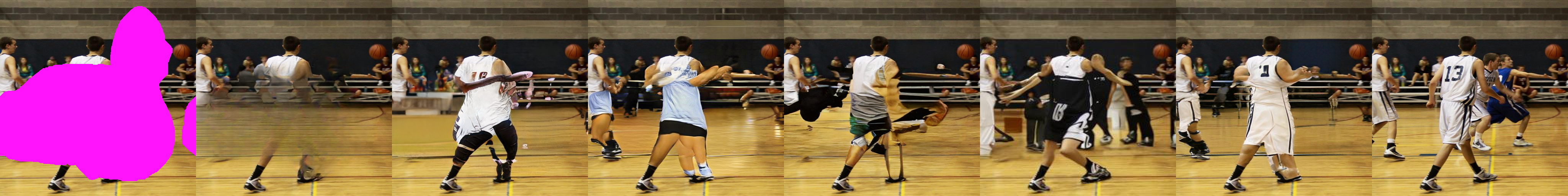}\\
	\includegraphics[width=1.\linewidth]{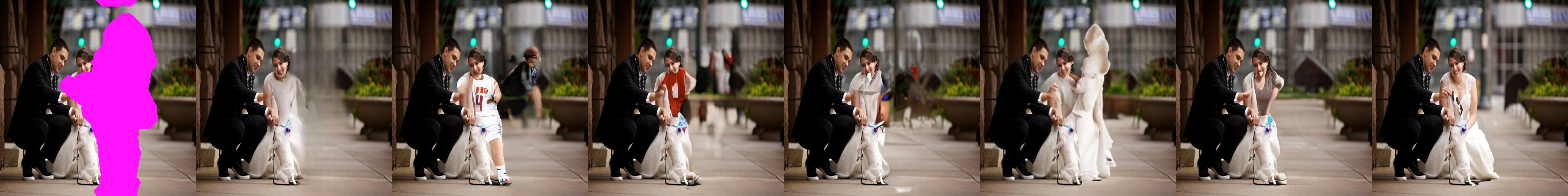}\\
	\includegraphics[width=1.\linewidth]{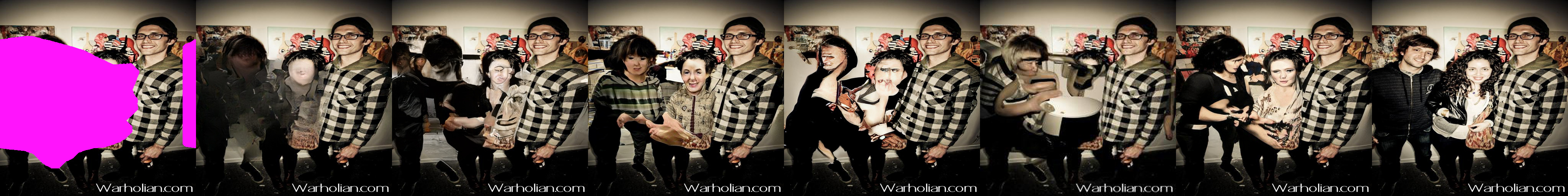}\\
	\includegraphics[width=1.\linewidth]{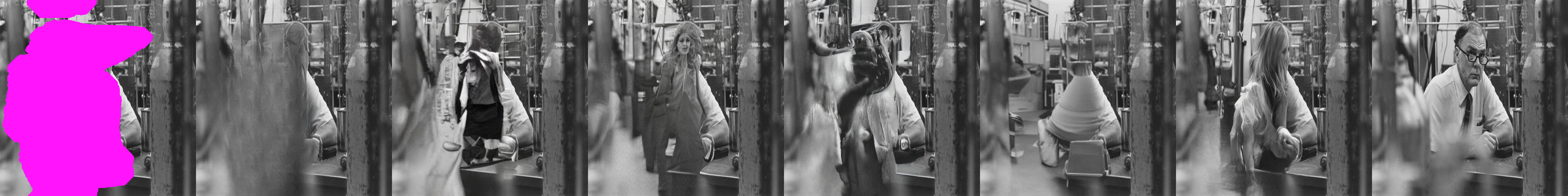}\\
	\includegraphics[width=1.\linewidth]{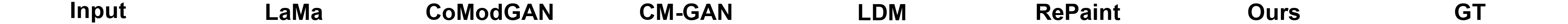}
	\caption{
		Qualitative comparisons on the standard inpainting task. Compared to the existing methods, our method can generate high-quality and photo-realist object instances.
	}
	\label{fig:supp_pure_inpainting}
\end{figure*}

\cref{fig:supp_main_panoptic} and \cref{fig:supp_main_panoptic_object} present the visual comparisons of various instance segmentation guided inpainting models. In addition, \cref{fig:supp_manipulation} presents more image manipulation examples generated by our instance segmentation guided inpainting model. \cref{fig:supp_semantic_guided_inpainting}, \cref{fig:supp_edge_guided_inpainting} and \cref{fig:supp_pure_inpainting} present the additional visual results of our approach on semantic-guided, edge-guided and standard inpainting tasks, respectively. \cref{fig:supp_pure_inpainting} includes visual comparisons to existing inpainting methods including Latent Diffusion~\cite{ldm} and RePaint~\cite{lugmayr2022repaint}.

\begin{table}[h]
	\caption{
		Quantitative evaluation on a standard inpainting task on a subset of Places2-person dataset that contains 4k samples. Our fully automatic pipeline, i.e. \emph{ours (automatic)}, achieves better numerical results.
		}
	\small
	\centering
		\begin{tabular}{l|c|c|c}
			\toprule
			Methods & FID$\downarrow$ & U-IDS (\%)$\uparrow$ & P-IDS (\%)$\uparrow$ \\
			\hline
			\multicolumn{4}{l}{\B \em CoModGAN masks}\\
			\hline
			LaMa~\cite{lama} & 33.1421 & 5.60 & 1.39 \\
			CoModGAN~\cite{comodgan} & 24.3939 & 12.65 & 4.74 \\
			CM-GAN~\cite{cmgan} & 20.4183 & 17.35 & 6.93 \\ 
			Latent Diffusion~\cite{ldm} & 24.8641 & 11.72 & 3.74\\ 
            RePaint~\cite{lugmayr2022repaint} & 30.6417 & 0.45 & 0.21 \\  
			\B ours (automatic) & \B 16.1032 & \B 22.40 & \B 12.13 \\ 	

			\hline
			\multicolumn{4}{l}{\B \em Object masks}\\
			\hline
			LaMa~\cite{lama} & 24.5222  & 8.46 & 1.66 \\
			CoModGAN~\cite{comodgan} & 20.3095 & 17.58 & 6.25\\
			CM-GAN~\cite{cmgan} & 17.4178 & 21.84 & 10.82 \\ 
			Latent Diffusion~\cite{ldm} & 19.7228 & 16.83 & 6.45\\
			RePaint~\cite{lugmayr2022repaint} & 22.4075 & 3.37 & 0.42\\
			\B ours (automatic) & \B 14.0794 & \B 26.05 & \B 14.21 \\  
			\bottomrule
		\end{tabular}
	\label{tab:supp_main_inpainting}
\end{table}


\section{Analysis of the LPIPS scores~\cite{lpips}}
\begin{figure*}[h]
	\centering
	\includegraphics[width=1\linewidth]{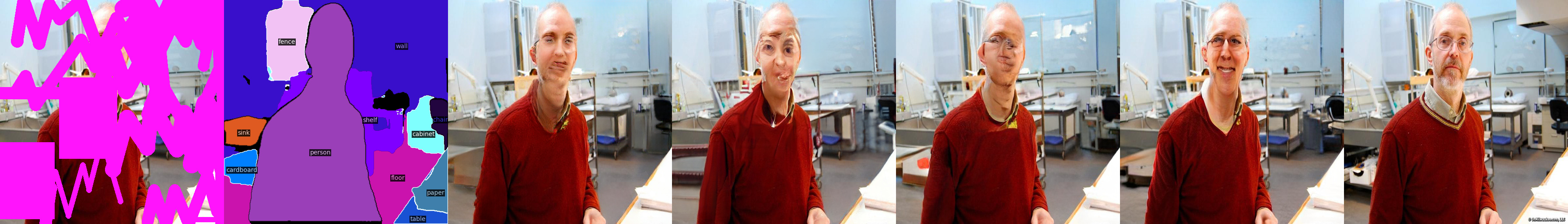}\\
	\includegraphics[width=1\linewidth]{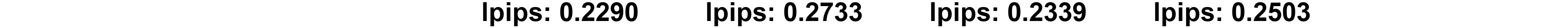}\\
	\includegraphics[width=1\linewidth]{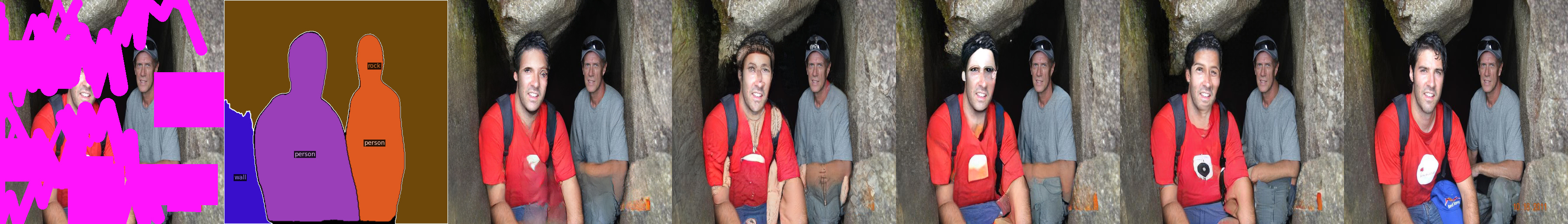}\\
	\includegraphics[width=1\linewidth]{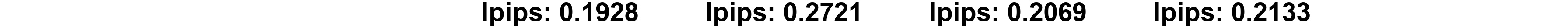}\\
	\includegraphics[width=1\linewidth]{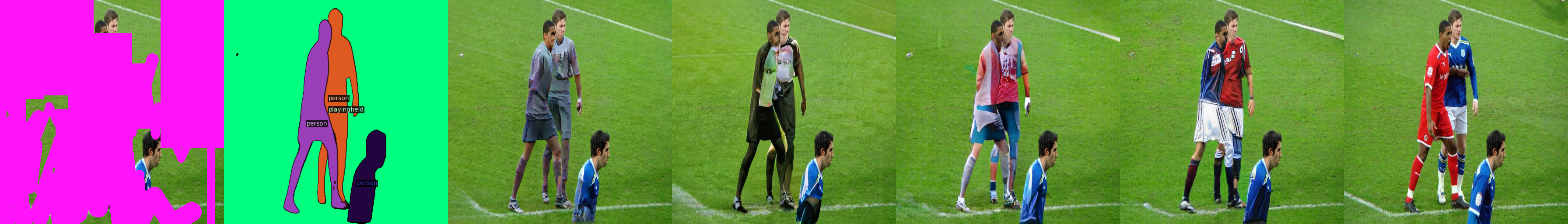}\\
	\includegraphics[width=1\linewidth]{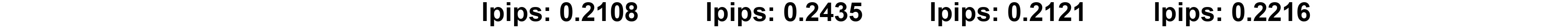}\\
	\includegraphics[width=1\linewidth]{figures_jpg/_vis_main_compare_panoptic_guided_inpainting/_vis_main_compare_panoptic_guided_inpainting_title_0.jpg}
	\caption{
		The Learned Perceptual Image Patch Similarity (LPIPS) metric~\cite{lpips} on individual images generated by LaMa*~\cite{lama}, CoModGAN*~\cite{comodgan}, CM-GAN*~\cite{cmgan}, and our approach on the instance map guided inpainting task. We found LPIPS favors averaged and overly-smooth outputs with faded structure details rather than results with realistic instances. Best viewed by zoom-in on screen.
	}
	\label{fig:supp_lpips}
\end{figure*}

We provide analysis on the LPIPS scores~\cite{lpips} on the images respectively generated by LaMa*~\cite{lama}, CoModGAN*~\cite{comodgan}, CM-GAN*~\cite{cmgan}, and our approach, in \cref{fig:supp_lpips}. We found that LPIPS is not a good metric for indicating the object-level realism as LPIPS tends to prefer faded out structures and give higher distance prediction to image completion results with better object-level realism such as face and body. Therefore, we do not evaluate the LPIPS metric in our further experiments.


\printbibliography

\begin{IEEEbiography}[{\includegraphics[width=1in,height=1.25in,clip,keepaspectratio]{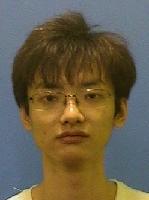}}]{Haitian Zheng} is a research scientist at Adobe Research. He received his Ph.D. degree in Computer Science from University of Rochester in May 2023, under the supervision of Prof. Jiebo Luo. Prior to that, he obtained his B.Sc. and the M.Sc. degrees in electronics engineering and informatics science from the University of Science and Technology of China, under the supervision of Prof. Lu Fang, in 2012 and 2016, respectively. His research interests include image generation, manipulation, inpainting and enhancement.
\end{IEEEbiography}

\begin{IEEEbiography}[{\includegraphics[width=0.95in,height=1.25in,clip,keepaspectratio]{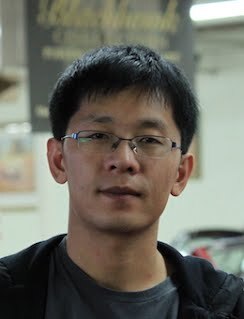}}]{Zhe Lin} is Senior Principal Scientist in Creative Intelligence Lab, Adobe Research. He received his Ph.D. degree in Electrical and Computer Engineering from University of Maryland at College Park in May 2009. Prior to that, he obtained his M.S. degree in Electrical Engineering and Computer Science from Korea Advanced Institute of Science and Technology in August 2004, and B.Eng. degree in Automation from University of Science and Technology of China. He has been a member of Adobe Research since May 2009. His research interests include computer vision, image processing, machine learning, deep learning, artificial intelligence. He has served as a reviewer for many computer vision conferences and journals since 2009, and recently served as an Area Chair for WACV 2018, CVPR 2019, ICCV 2019, CVPR 2020, ECCV 2020, ACM Multimedia 2020.
\end{IEEEbiography}

\begin{IEEEbiography}[{\includegraphics[width=0.95in,height=1.25in,clip,keepaspectratio]{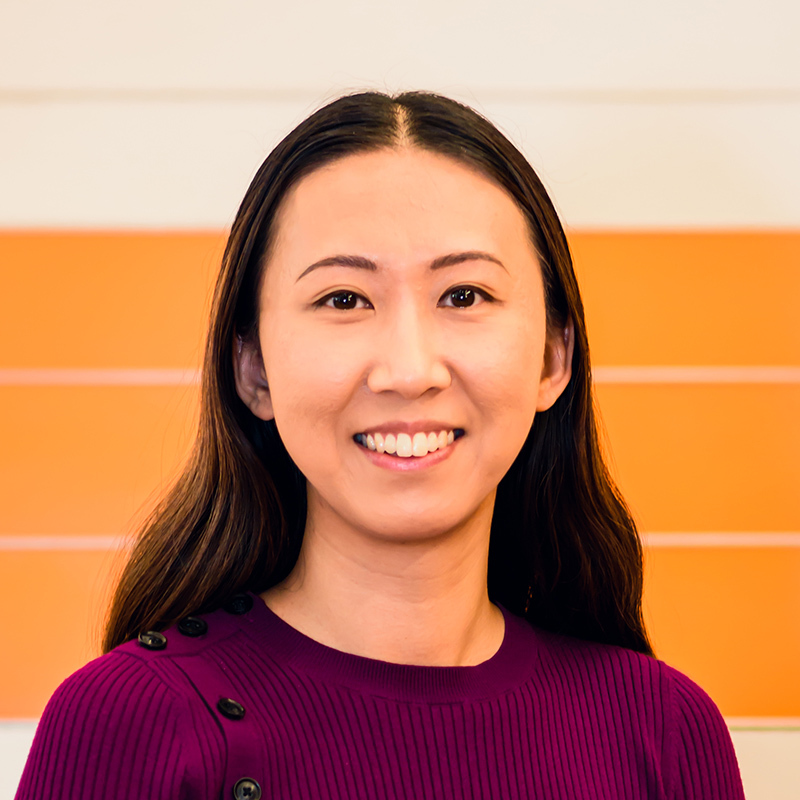}}]{Jingwan Lu} joined Adobe Research in 2014. She is currently leading a team of research scientists and engineers with the vision to disrupt digital imaging and design with big data and generative AI. Her research interests include generative image modeling (GANs, etc.), computational photography, digital human and data-driven artistic content creation. She has over 20 issued patents and over 40 publications in top vision, graphics and machine learning conferences. Some of them attracted lots of public attentions for example, Scribbler, VoCo, StyLit, FaceStyle, Playful Palette. Jingwan served as a program committee member for Siggraph, Siggraph Asia and Eurographics.
\end{IEEEbiography}

\begin{IEEEbiography}[{\includegraphics[width=1in,height=1.25in,clip,keepaspectratio]{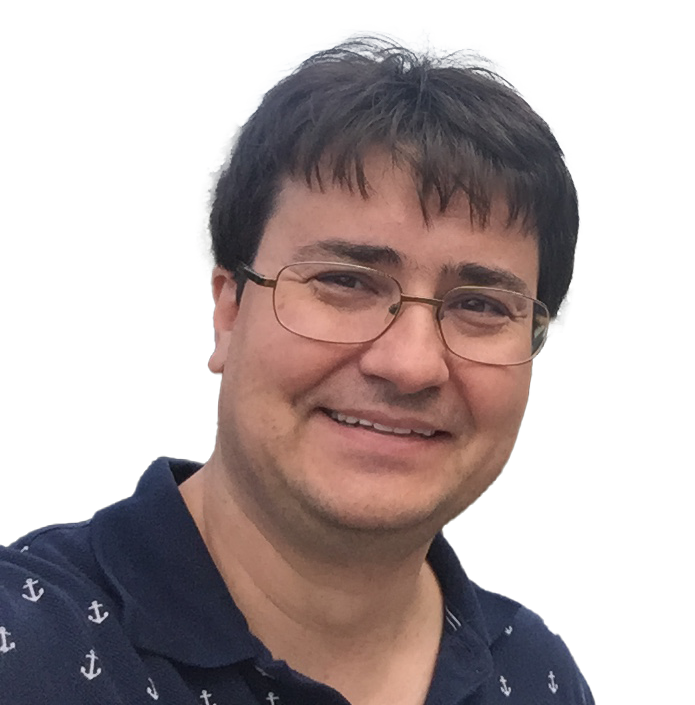}}]{Scott Cohen}
received the B.S. degree in mathematics from Stanford University, Stanford, CA, in 1993, and the B.S., M.S., and Ph.D. degrees in computer science from Stanford University in 1993, 1996, and 1999, respectively. He is currently a Senior Principal Scientist and manager of a Computer Vision research team at Adobe Research, San Jose, CA. His research interests include image segmentation, editing, and understanding such as object attribute prediction, visual grounding, captioning, and search.
\end{IEEEbiography}

\begin{IEEEbiography}[{\includegraphics[width=1in,height=1.25in,clip,keepaspectratio]{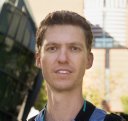}}]{Eli~Shechtman}
is a Senior Principal Scientist at the Creative Intelligence Lab, Adobe Research. He received his B.Sc. in Electrical Engineering (cum laude) from Tel-Aviv University in 1996 and his MSc and PhD with honors in Applied Mathematics and Computer Science from the Weizmann Institute of Science in 2003 and 2007. 
He then joined Adobe and also shared his time as a post-doc at the University of Washington between 2007-2010. He has published over 100 academic publications. Two of his papers were chosen to be published as “Research Highlight” papers in the Communication of the ACM (CACM) journal. He served as a Technical Paper Committee member at SIGGRAPH 2013 and 2014, was an Area Chair at ICCV 2015, 2019 and 2021, CVPR 2015, 2017 and 2020, 2023, ECCV 2022 and an Associate Editor for TPAMI from 2016-2020. He has received several honors and awards, including the Best Paper prize at ECCV 2002, a Best Paper award at WACV 2018, a Best Paper Runner Up at FG 2020 and the Helmholtz “Test of Time Award” at ICCV 2017. His research is in the intersection of computer vision, computer graphics and machine learning. 
\end{IEEEbiography}

\begin{IEEEbiography}[{\includegraphics[width=1in,height=1.25in,clip,keepaspectratio]{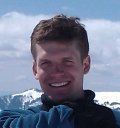}}]{Connelly~Barnes} 
is a Senior Research Scientist at the Creative Intelligence Lab, Adobe Research. He received his Ph.D. degree in Computer Science from Princeton University in May 2011. Prior to that, he obtained his honor B.S. degrees in Mathematics and Computational Physics in Oregon State University.
His career has focused on computer graphics and vision topics, such as image and video processing, texture, deep learning for image tasks, and domain specific languages especially for shaders. 
He have served on these technical papers committees: CGDIP (2017), Eurographics short papers (2013), Eurographics Symposium on Rendering (2013, 2014, 2017), ICCP (2013), Pacific Graphics (2017), SIGGRAPH (2015, 2016, 2019, 2020), SIGGRAPH Asia (2018, 2021).
\end{IEEEbiography}

\begin{IEEEbiography}[{\includegraphics[width=1in,height=1.25in,clip,keepaspectratio]{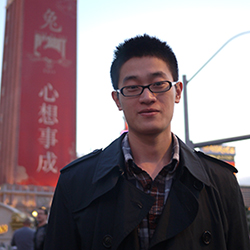}}]{Jianming Zhang}
is a senior research scientist at Adobe. His research interests include visual saliency, image segmentation, 3D understanding from a single image and image editing. He got his PhD degree in computer science at Boston University in 2016.
\end{IEEEbiography}

\begin{IEEEbiography}[{\includegraphics[width=1in,height=1.25in,clip,keepaspectratio]{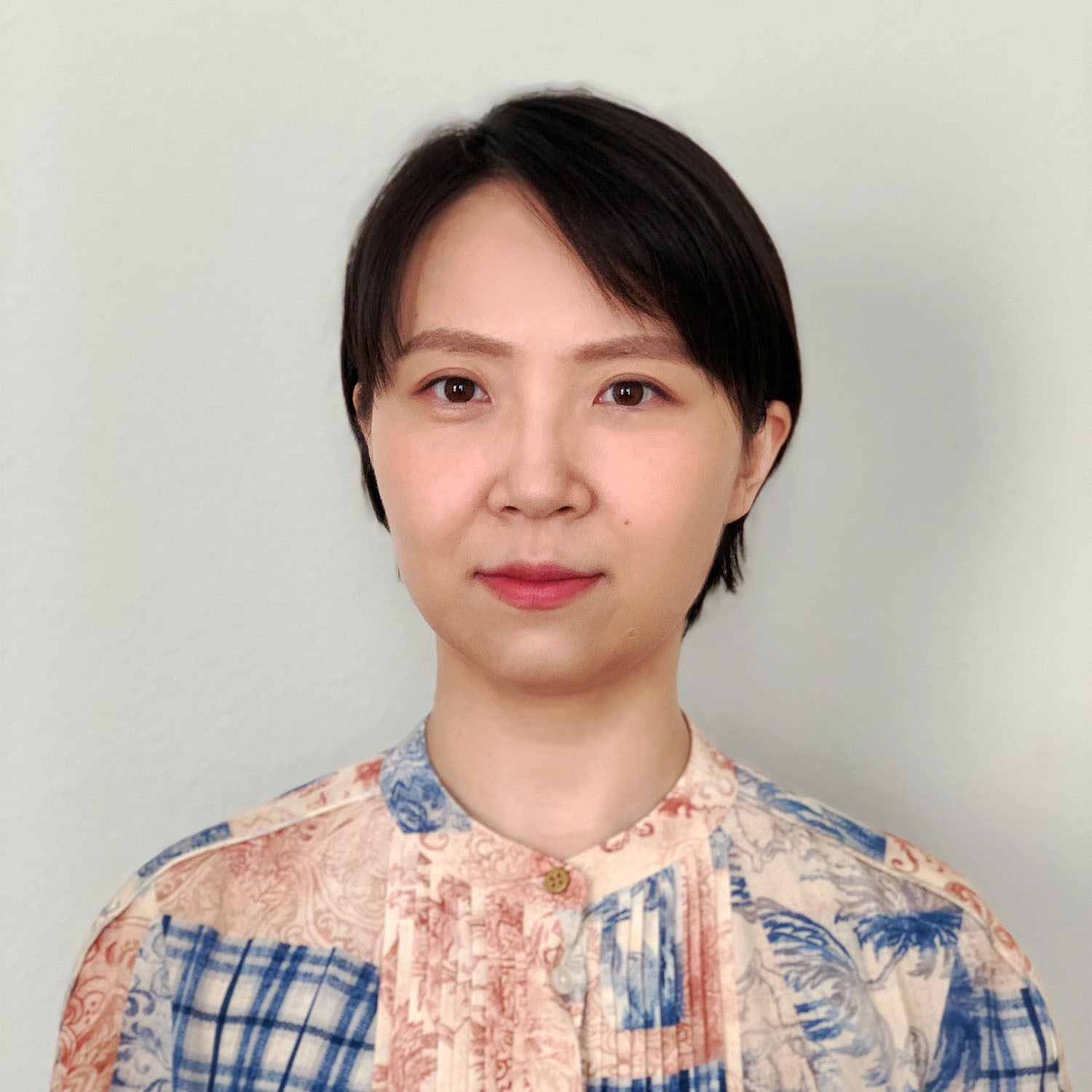}}]{Qing~Liu}
is a research scientist/engineer at Adobe. Her research interests include generative models, image understanding, image editing, and weakly supervised learning. She received her Ph.D. degree in Computer Science from Johns Hopkins University in 2022.
\end{IEEEbiography}

\begin{IEEEbiography}[{\includegraphics[width=1in,height=1.25in,clip,keepaspectratio]{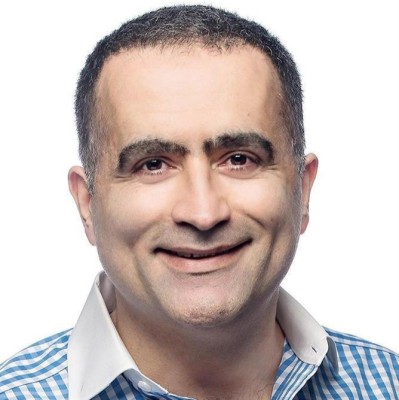}}]{Sohrab~Amirghodsi}
is a Principal Computer Scientist at Adobe Inc. He received his Master and B.S. degree in Computer Science from University of Washington. He leads Photoshop AI image Cleanup Lab. In the current role, he is responsible for defining the AI roadmaps for his lab. He participates in research and development of various complex AI pipelines at Adobe.
\end{IEEEbiography}

\begin{IEEEbiography}[{\includegraphics[width=1in,height=1.25in,clip,keepaspectratio]{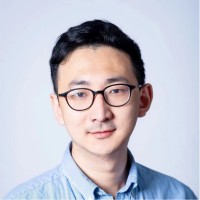}}]{Yuqian~Zhou}
is a research scientist at Adobe. He received his Ph.D. from UIUC and was advised by Prof. Mark Hasegawa-Johnson. Before that, he was supervised by Prof. Thomas Huang(1936-2020). He was a research assistent at Image Formation and Processing Group IFP led by Prof. Humphrey Shi of the University of Illinois at Urbana-Champaign (UIUC). He received his Bachelor and Mphil. degree from the Hong Kong University of Science and Technology(HKUST), working at Neuromorphic Interactive System Lab (NISL) supervised by Prof. Bertram Shi.
\end{IEEEbiography}

\begin{IEEEbiography}[{\includegraphics[width=1in,height=1.25in,clip,keepaspectratio]{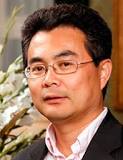}}]{Jiebo Luo}
Jiebo Luo (S93, M96, SM99, F09) is a Professor of Computer Science at the University of Rochester since 2011 after a prolific career of over 15 years with Kodak Research. He has authored over 500 technical papers and holds over 90 U.S. patents. His research interests include computer vision, NLP, machine learning, data mining, computational social science and digital health. He has served as a Program Co-Chair of the ACM Multimedia 2010, IEEE CVPR 2012, ACM ICMR 2016, and IEEE ICIP 2017, and a General Co-Chair of ACM Multimedia 2018. He has served on the Editorial Boards of the IEEE
TRANSACTIONS ON PATTERN ANALYSIS AND MACHINE INTELLIGENCE, IEEE TRANSACTIONS ON MULTIMEDIA, IEEE TRANSACTIONS ON CIRCUITS AND SYSTEMS FOR VIDEO TECHNOLOGY,  IEEE TRANSACTIONS ON BIG DATA,  Pattern Recognition, Machine Vision and Applications, and ACM Transactions on Intelligent Systems and Technology. He is the Editor-in-Chief of the IEEE TRANSACTIONS ON MULTIMEDIA for 2020-2022. Professor Luo is also a Fellow of ACM, AAAI, SPIE and IAPR.
\end{IEEEbiography}





\end{document}